\pdfoutput=1
\documentclass[11pt]{article}

\usepackage[final]{acl}

\usepackage{times}
\usepackage{latexsym}
\usepackage{subfigure}
\usepackage{amssymb}
\usepackage{utfsym}
\usepackage{pifont}
\usepackage{makecell}
\usepackage{colortbl}
\usepackage[T1]{fontenc}

\usepackage[utf8]{inputenc}

\usepackage{microtype}

\usepackage{inconsolata}

\usepackage{graphicx}
\newcommand{\ourmethod}{{\fontfamily{ppl}\selectfont Med-RewardBench}}

\title{\ourmethod: Benchmarking Reward  Models and Judges \\ for Medical Multimodal Large Language Models}

\author{
  Meidan Ding$^{1,2,3,\dag}$, 
  Jipeng Zhang$^{4,\dag}$, 
  Wenxuan Wang$^{5}$, 
  Cheng-Yi Li$^{6}$,\\
  \textbf{Wei-Chieh Fang}$^{7}$, 
  \textbf{Hsin-Yu Wu}$^{6}$, 
  \textbf{Haiqin Zhong}$^{8}$,
  \textbf{Wenting Chen}$^{9}$, 
  \textbf{Linlin Shen}$^{1,2,3}$ \\
  $^{1}$College of Computer Science and Software Engineering, Shenzhen University \\
  $^{2}$School of Artificial Intelligence, Shenzhen University
  $^{9}$City University of Hong Kong \\
  $^{3}$Guangdong Provincial Key Laboratory of Intelligent Information Processing \\
  $^{4}$The Hong Kong University of Science and Technology 
  $^{5}$Renmin University of China\\
  $^{6}$National Yang Ming Chiao Tung University
  $^{7}$Taipei Veterans General Hospital \\
  $^{8}$School of Biomedical Engineering, Shenzhen University
}

\begin{document}
\maketitle
\newcommand\blfootnote[1]{%
\begingroup
\renewcommand\thefootnote{}\footnote{#1}%
\addtocounter{footnote}{-1}%
\endgroup
}

\blfootnote{\textsuperscript{\dag} These authors contributed equally.}

\begin{abstract}

Multimodal large language models (MLLMs) hold significant potential in medical applications, including disease diagnosis and clinical decision-making. However, these tasks require highly accurate, context-sensitive, and professionally aligned responses, making reliable reward models and judges critical. Despite their importance, medical reward models (MRMs) and judges remain underexplored, with no dedicated benchmarks addressing clinical requirements. Existing benchmarks focus on general MLLM capabilities or evaluate models as solvers, neglecting essential evaluation dimensions like diagnostic accuracy and clinical relevance.
To address this, we introduce \textbf{\ourmethod}, the first benchmark specifically designed to evaluate MRMs and judges in medical scenarios. \ourmethod~features a multimodal dataset spanning 13 organ systems and 8 clinical departments, with 1,026 expert-annotated cases. A rigorous three-step process ensures high-quality evaluation data across \textit{six clinically critical dimensions}. We evaluate 32 state-of-the-art MLLMs, including open-source, proprietary, and medical-specific models, revealing substantial challenges in aligning outputs with expert judgment. Additionally, we develop baseline models that demonstrate substantial performance improvements through fine-tuning.
This work provides a foundation for improving and evaluating reward models and judges in medical AI, which could speed up the creation of more reliable and practical MLLMs.
Source code and data are to be released.
\end{abstract}

\section{Introduction}
\begin{figure}[t]
    \centering
    \includegraphics[width=\columnwidth]{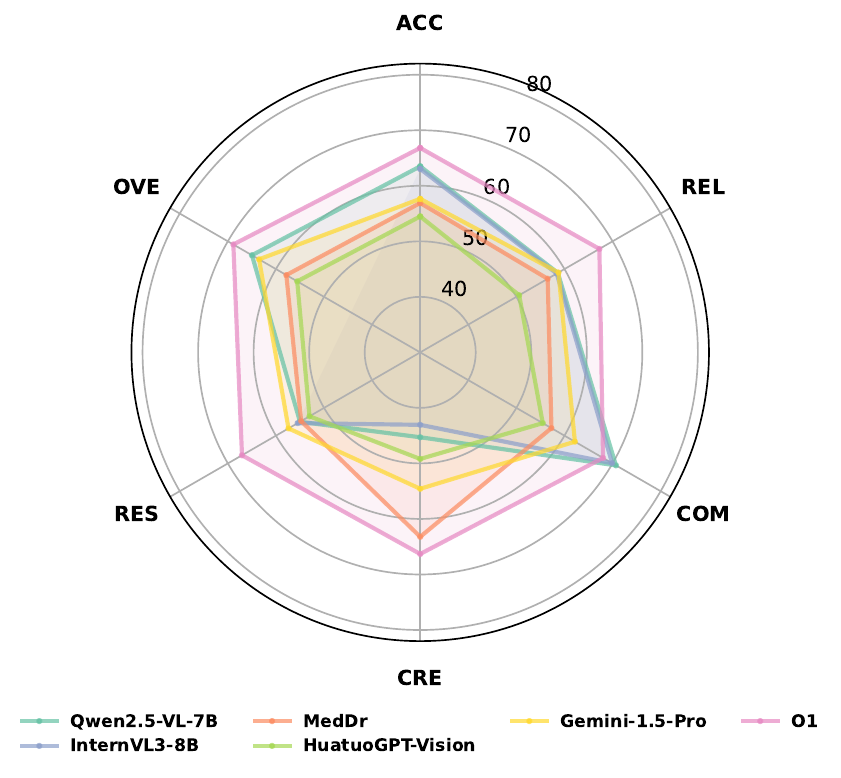}
    \caption{Radar chart showing various MLLMs' performance on~\ourmethod~across organ systems in six key dimensions: Accuracy (ACC), Relevance (REL), Comprehensiveness (COM), Creativity (CRE), Responsiveness (RES), and Overall (OVE).}
    \label{fig:intro}
\end{figure}

Multimodal large language models (MLLMs) have shown significant potential in supporting disease diagnosis~\cite{liu2023medical}, clinical decisions, and treatment recommendations~\cite{moglia2024minigpt} with a unified framework. This progress is largely driven by the rapid advances in MLLMs~\cite{zhu2023minigpt,liu2024llava} and the increasing availability of structured medical data~\cite{li2024llavamed}. However, unlike general scenarios where responses can tolerate a certain degree of creativity or ambiguity, medical scenes strongly demand responses that are highly accurate, context-sensitive, and aligned with professional medical standards due to the potential consequences of misinformation.


In order to satisfy such requirements, researchers focused on developing automatic verification mechanisms to guide and constrain MLLM behavior, ensuring that the outputs reflect expert-level reasoning. Such verification mechanisms, typically formulated as reward models or judges, are crucial for the development of MLLMs. High-quality Multimodal Reward Models (MRMs)~\cite{xiong2024llava,chen2025rm} and MLLM-as-a-judge~\cite{chen2024mllm,pu2025judge} serve critical functions during different stages for MLLMs: during training, they can provide rule-based reward for reinforcement learning, directly influencing stability and outcomes~\cite{ouyang2022training,sun2023aligning,zhang2025r1}; during inference, MRMs and judges facilitate test-time scaling strategies, such as best-of-n selection of responses~\cite{wang2025visualprm,zhou2025evaluating}; and in evaluation contexts, MRMs and judges can function as automated evaluators, particularly for open scenarios~\cite{xiong2024llavacritic,laskar2025judging}.

\noindent \textbf{Medical Multimodal Reward Models or Judges Are Underexplored.} Despite advances, MRMs or judges remain underexplored in the medical domain, with no standardized benchmark available for their evaluation. Existing medical benchmarks are proposed to evaluate in specialized medical fields~\cite{he2020pathvqa,SurgicalVQA,kvasir-vqa,bae2024mimic} and general medical scenario~\cite{zhang2023pmc,hu2024omnimedvqa,li2024gmai,zuo2025medxpertqa,ben2019vqa,liu2021slake}, primarily assessing MLLM on specific tasks such as predicting diagnosis or recommending treatment. However, these benchmarks are designed to evaluate model capabilities as solvers, not as judges, making them inadequate to evaluate Med-MLLM's response quality. Therefore, a dedicated medical reward benchmark is essential to properly evaluate reward models in the medical domain.

\noindent \textbf{General Reward Benchmarks Miss Critical Dimensions in Clinical Evaluation.} Recently, some general MLLM reward benchmarks~\cite{li2024vlrewardbench,pu2025judge,xiong2024llavacritic} have emerged to evaluate MLLMs' judgment capabilities. These benchmarks typically span diverse domains (e.g. math, animal) and assess reward models across multiple dimensions, including instruction-following, hallucination detection, and reasoning ability~\cite{ruan2025vlrmbench,son2024mm}. Although these benchmarks provide valuable insights across general domains, they do not consider medical scenarios or the specific requirements of clinical decision-making. For the computer-aided algorithms, clinicians prioritize several dimensions of generated responses, such as diagnostic accuracy, clinical relevance, and evidence-based responsiveness~\cite{yang2025application}. These critical aspects are not considered by the current general reward benchmarks.

\begin{table*}[t]
\centering
\caption{Existing reward model benchmarks for various domains, including specialized medical fields, general medical scenario, and general-purpose scenario (general), compared in expert annotation, pairwise evaluation, judgment capability evaluation, and multi-dimensional evaluation. }
\scalebox{0.75}{
\begin{tabular}{lcccccc}
\hline
\textbf{Benchmark} &\textbf{Size} & \textbf{Domain} & \textbf{Expert Annotation} & \textbf{Pairwise} & \textbf{Judgement} & \textbf{Multi-dimension}   \\
\hline
\rowcolor{green!10}
VQA-RAD~\cite{lau2018dataset} &451            & Radiology  & \ding{55}  & \ding{55}     & \ding{55} & \ding{55}  \\
\rowcolor{green!10}
VQA-Med~\cite{ben2019vqa}   &500          & Radiology  & \ding{55}  & \ding{55}     & \ding{55} & \ding{55}  \\
\rowcolor{green!10}
Path-VQA~\cite{he2020pathvqa}  &6,719          & Pathology  & \ding{55}  & \ding{55}     & \ding{55} & \ding{55}  \\
\rowcolor{green!10}
SLAKE~\cite{liu2021slake}   &1,061            & Radiology  & \ding{55}  & \ding{55}     & \ding{55} & \ding{55}  \\
\rowcolor{yellow!10}
PMC-VQA~\cite{zhang2023pmc}   &2,000          & Medical    & \ding{55}  & \ding{55}     & \ding{55} & \ding{55} \\
\rowcolor{yellow!10}
OmniMedVQA~\cite{hu2024omnimedvqa}  &127,995        & Medical    & \ding{55}  & \ding{55}     & \ding{55} & \ding{55}  \\
\rowcolor{yellow!10}
GMAI-MMBench~\cite{ye2024gmai} &21,281       & Medical    & \ding{55}  & \ding{55}     & \ding{55} & \ding{55}  \\
\rowcolor{yellow!10}
MMMU(H\&M)~\cite{yue2024mmmu}  &1,752        & Medical    & \checkmark      & \ding{55}     & \ding{55} & \ding{55}  \\
\rowcolor{yellow!10}
MMMU-Pro(H\&M)~\cite{yue2024mmmu-pro} &346     & Medical    & \checkmark      & \ding{55}    & \ding{55} & \ding{55}  \\
\rowcolor{yellow!10}
MedXpert MM~\cite{zuo2025medxpertqa}   &2,000      & Medical    & \checkmark      & \ding{55}     & \ding{55} & \ding{55}  \\
\rowcolor{orange!10}
VL-RewardBench~\cite{li2024vlrewardbench} &1,250     & General    & \ding{55}  & \checkmark & \checkmark & \ding{55}  \\ \hline
\textbf{Med-RewardBench} &1,026 & Medical & \checkmark     & \checkmark    & \checkmark & \checkmark  \\
\hline
\end{tabular}}
\label{tab:benchmark-comparison}
\end{table*}


Therefore, we need a medical-specific benchmark that can evaluate the capabilities of the reward model with multiple dimensions critical for clinical applications.

To address these challenges, we introduce \textbf{\ourmethod}, a novel benchmark designed to evaluate the judgment capabilities of MLLMs as reward models in medical scenarios. \ourmethod~is built upon a comprehensive multimodal medical evaluation dataset that spans 13 organ systems and 8 clinical departments, consisting of 1,026 cases with expert annotations. The construction of \ourmethod~follows a rigorous three-step process. 1) \textit{Image-question pair collection}: We curate data from five diverse datasets covering various tasks. Using five MLLMs, we generate responses to each question, identify high-quality questions that fewer than three MLLMs can answer correctly, and have clinicians rigorously assess them across several criteria. 2) \textit{MLLM response collection}: For each image-question pair in \ourmethod, we employ 12 MLLMs to generate responses. Two responses are uniformly sampled from these outputs as choices for reward model evaluation. The resulting instruction data consists of an image-question pair and two response choices. 3) \textit{Comparison with human annotations}: We recruit 3 general practitioners to annotate the instruction data across \textit{6 dimensions critical for clinical application}, with consistency verified through majority voting. Then, we evaluate 32 state-of-the-art MLLMs on \ourmethod, ranging from open-source models (3B-72B), medical-specific models to proprietary models like GPT-4o and O1. Our findings reveal significant challenges for current models in Fig~\ref{fig:intro}. Even the most advanced proprietary models achieve moderate performance, while medical-specific models like HuatuoGPT-Vision struggle to perform better than random chance. Furthermore, we establish baseline models on our training dataset and demonstrate substantial performance improvements through fine-tuning.
Our contributions can be summarized as follows: 

\begin{itemize}
    \item We introduce \textbf{\ourmethod}, the first comprehensive benchmark designed to evaluate reward models and judges in medical scenarios, uniquely integrating multimodal medical data across 13 organs and 8 departments, and establishing a standardized methodology for assessing MLLMs' performance in six key dimensions.  
    \item  We propose a comprehensive evaluation framework that encompasses diverse MLLM categories, including open-source models, proprietary models, and specific medical MLLMs. Our methodology provides a quantitative assessment of model performance through careful measurement of alignment between MLLM outputs and human expert judgment, establishing a comprehensive framework for comparing model capabilities across different clinical domains and specialties.
    \item We develop baseline models on our curated training dataset. The results show a substantial improvement in judgment capabilities. 
\end{itemize}

\section{Related Work}

\subsection{Medical Multimodal Benchmarks}

Traditional medical multimodal benchmarks can be broadly categorized into two types: specific and general-purpose. Specific benchmarks focus on a specific modality or medical domain. VQA-RAD ~\cite{lau2018dataset}, VQA-Med~\cite{ben2019vqa}, and SLAKE ~\cite{liu2021slake} are primarily centered on radiology, while Path-VQA ~\cite{he2020pathvqa} focuses on pathology. These benchmarks provide extensive evaluation for their intended specialties, yet have a highly constrained scope and limited generalizability. With advancements in MLLMs, recent developments in general-purpose benchmarks, like PMC-VQA~\cite{zhang2023pmc}, OmniMedVQA~\cite{hu2024omnimedvqa}, and GMAI-MMBench~\cite{ye2024gmai}, provide more comprehensive evaluations.  MMMU (H\&M) Series ~\cite{yue2024mmmu,yue2024mmmu-pro} and MedXpertQA~\cite{zuo2025medxpertqa} encompass a wider variety of diagnostic scenarios.
However, they are designed to evaluate model capabilities as solvers, not as judges. This work breaks this barrier by proposing a medical reward benchmark, which evaluates models from the perspective of judgment and alignment with expert preferences, rather than mere task resolution.  

\begin{figure*}[t]
    \centering
    \includegraphics[width=\textwidth]{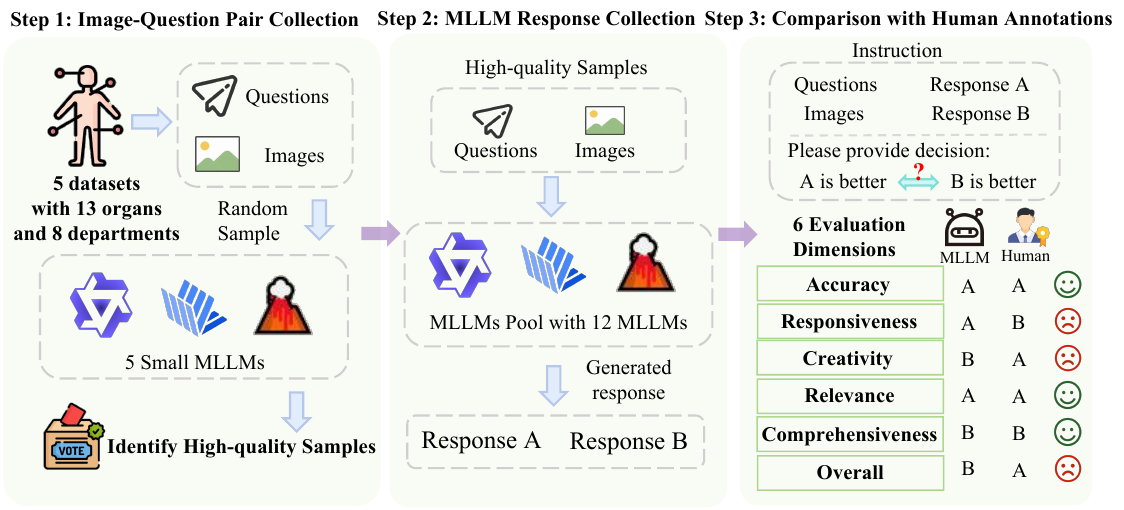}
    \caption{Overview of \textbf{\ourmethod}~with three-step construction process, i.e., 1) image-question pair collection, 2) MLLM response collection, and 3) comparison with human annotations. }
    \label{fig:method}
\end{figure*}

\subsection{Benchmarks for Reward Models and Judges}
The evolution of multimodal large language models (MLLMs) and multimodal reward models (MRMs) has significantly enhanced their capabilities as evaluators across a wide range of general tasks. Recent studies have demonstrated the potential of MLLMs and MRMs in assessing performance and aligning with human preferences. For instance, \citeauthor{chen2024mllm} proposed a benchmark termed MLLM-as-a-Judge, which evaluates the judge capabilities of MLLMs across diverse modalities, including text, images, and audio. \citeauthor{pu2025judge} extends MLLM-as-a-Judge across modalities to a unified manner by introducing two benchmarks, TASKANYTHING and JUDGEANYTHING, to respectively evaluate the overall performance and judging capabilities of MLLMs across
any-to-any modality tasks.
Additionally, \citeauthor{li2024vlrewardbench} introduced VL-RewardBench, a comprehensive benchmark designed to evaluate MRMs across general queries, visual hallucination detection, and complex reasoning tasks. 
While these benchmarks have significantly advanced the evaluation of judges and MRMs in general domains, there remains a notable gap in the development of specific reward benchmarks for medical tasks.  Thus, we propose \ourmethod~to evaluate the ability of judges and reward models in medical multimodal tasks with six key dimensions.

\section{\ourmethod}

In this section, we introduce the construction process of \ourmethod~in Fig.~\ref{fig:method}. Specifically, \ourmethod~consists of preference pairs $(X,I,R_a,R_r)$, where $X$ represents the input image, $I$ denotes a user instruction, and $(R_a,R_r)$ denotes two different responses, respectively. For simplicity, we focus on single-image, single-turn interactions. In later subsections, we respectively describe the three steps of the construction process: 1) image-question pair collection, 2) MLLM response collection, and 3) comparison with human annotations.

\subsection{Step 1: Image-Question Pair Collection}
\label{step1_sec}
Our~\ourmethod~development prioritized comprehensive coverage of clinical scenarios through careful selection from five publicly available medical datasets, including PubMedVision~\cite{chen2024huatuogpt}, LLaVA-Med~\cite{li2024llavamed}, Quilt-Instruct~\cite{seyfioglu2024quilt}, CARES~\cite{xia2024cares}, and RULE~\cite{xia2024rule}. 
These datasets span multiple domains, including radiology, histology, and general medicine, encompassing diverse organ systems and task types such as diagnosis, localization, and descriptive analysis. The initial dataset, denoted as $P=\left \{(X_1,Q_1),...,(X_n,Q_n) \right \}$, comprising multiple pairs of medical images $X$ and corresponding questions $Q$.

To ensure the quality and complexity of the dataset, we implemented a systematic multi-step filtration process. The first phase employed five small MLLMs as weak evaluators to assess the difficulty of instruction, including DeepSeek-VL-1.3B-chat, Qwen2-VL-2B-Instruct, blip2-opt-2.7b, paligemma-3b-mix-224, and h2ovl-mississippi-2b. Then, we select the pairs that can be answered correctly by fewer than three MLLMs and include these "difficult" pairs in our~\ourmethod.

After the initial filtering, we implemented a balanced sampling strategy by randomly selecting eighty instructions per organ category to ensure equitable representation across medical domains. These selected instructions were rigorously evaluated by a panel of medical professionals who evaluated their clinical relevance, accuracy, complexity, and image quality. Instructions deemed ambiguous, irrelevant, or insufficiently complex were either refined or eliminated from the dataset. The culmination of this comprehensive filtering and verification process resulted in a high-quality dataset comprising 1026 image-question pairs with 13 organs and 8 departments. The number of pairs for each organ are displayed in Table~\ref{tab:overall}.

\subsection{Step 2: MLLM Response Collection}
To generate diverse responses based on image-question pairs, we employ a comprehensive MLLM pool with twelve widely used MLLMs. Our selection encompasses models ranging from 7 billion to 72 billion parameters, incorporating the most prominent architectures in the field. This diversity aims to create a rich response pool that captures a broad spectrum of reasoning patterns, linguistic styles, and potential error modes. For each image-question pair in our \ourmethod, we uniformly sampled two responses from the generated pool. This sampling strategy facilitates balanced and fair comparisons by ensuring each model has an equal probability of representation in the evaluation process. The resulting dataset, denoted as $D = \{(X_i,Q_i,R_i)|(X_i,Q_i)\in P\}$, where $R_i = \{R_a,R_r\}$ represents the pair of responses for each image-question pair, serves as the foundation for the subsequent evaluation phase. We carefully design the A/B choice ratio to ensure balance. Statistical analysis shows that option A is chosen as the correct answer in 51.3\% of cases, while option B is correct in 48.7\% of cases.

\subsection{Step 3: Comparison with Human Annotations}
For the expertise of the annotation, data annotation was conducted by 3 medical experts, all registered general practitioners with 4-5 years of clinical experience. To select the qualified experts, we established specific criteria, including clinical experience requirements, board certification status, and demonstrated expertise in general medicine. 
All selected experts completed standardized training with an annotation guideline (Appendix Fig.~\ref{fig:guideline}) on our annotation protocols before beginning the formal evaluation. The annotation focused on six evaluation dimensions. 1) \textbf{Accuracy}: The correctness of the medical information provided.
2) \textbf{Relevance}: The extent to which the response directly addresses the given instruction.
3) \textbf{Comprehensiveness}: The degree to which the response covers all relevant aspects of the question.
4) \textbf{Creativity}: The ability to offer insightful or innovative interpretations of the instruction.
5) \textbf{Responsiveness}: The model’s capacity to provide timely and appropriate feedback to patient-related inquiries.
6) \textbf{Overall}: A holistic assessment of the response’s quality and utility.
The instruction $I$ includes the system prompt $S$ and questions $Q$, image, and two responses (Appendix Fig.~\ref{fig:prompt}). For each instruction, experts selected the superior response across six dimensions. Uncertain cases were resolved through majority voting, consistent with standard medical annotation practices. The annotation process involved multiple rounds of review and refinement to ensure consistency and quality across the entire dataset. 

\begin{figure}[t]
    \centering
    \includegraphics[width=\columnwidth]{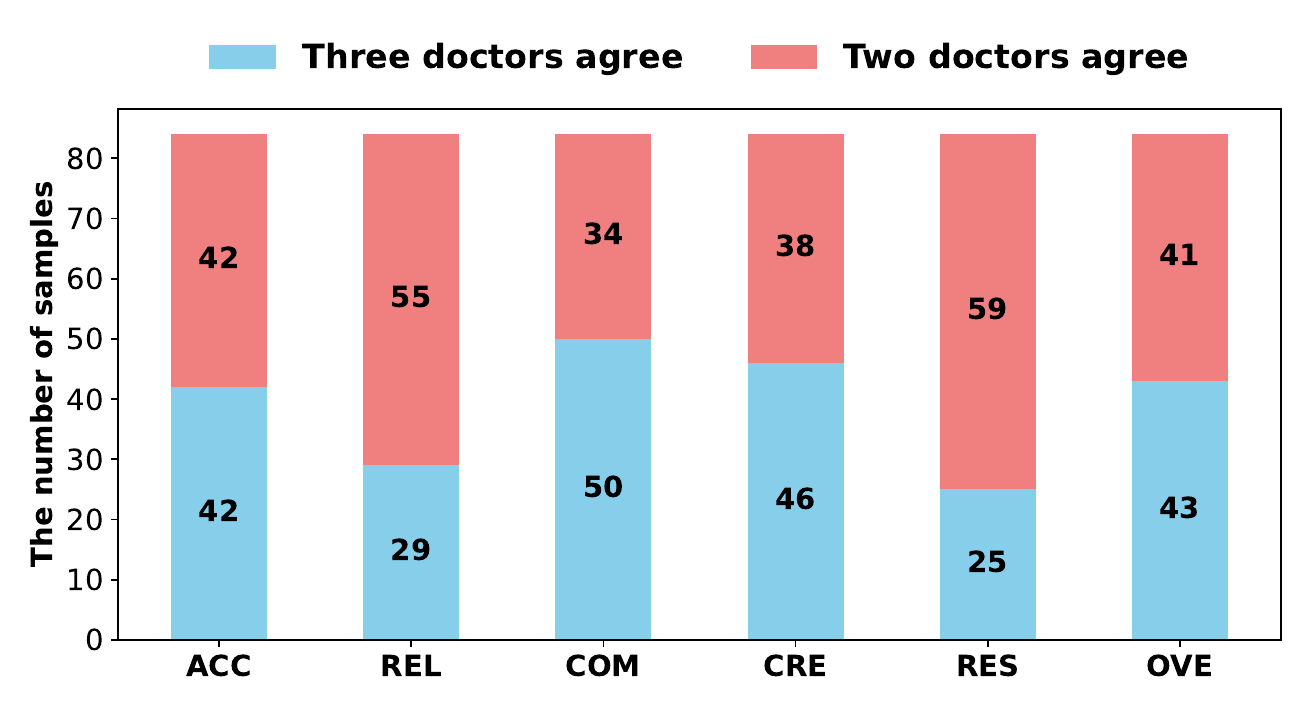}
    \caption{Consistency statistics of experts’ annotations.  }
    \label{fig:agreement}
\end{figure}

To validate the reliability of our benchmark, we randomly selected 84 data samples from the uncertain samples to verify the consistency among annotators. Fig.~\ref{fig:agreement} presents the level of agreement among experts, showing consensus achieved by at least 2 experts. Across all six dimensions, every answer received agreement from at least two experts. For most dimensions, all three experts achieved complete consensus.

Finally, we evaluated different MLLMs on~\ourmethod. Specifically, we calculated the overall consistency between model-selected preferences and human-selected preferences across all six dimensions.

\begin{table*}[t]
\centering
\footnotesize
\renewcommand\arraystretch{1.3}
\tabcolsep=0.1cm
\scalebox{0.75}{
\begin{tabular}{l|c|c|cccccccccccccc}
\hline
Model &Year & Size & ABD  & BRE  & BRN   & CHE  & EYE    & FOT   & GI  & HRT  & LL & LNG  & OC & PC & UL & Overall \\ 
&&&(79) &(80) &(80) &(80) &(73) &(80) &(78) &(80) &(80) &(80) &(76) &(80) &(80) &(1026) \\ \hline
\rowcolor{gray! 10}\multicolumn{17}{c}{Open-Source MLLMs} \\

VILA1.5-3B &2024 &3B &58.23  &69.64  &58.82  &57.89  &43.66  &62.75  &52.94 &52.00 &52.00  &52.00  &52.11  &59.26  &52.11  &55.64 \\
xGen-MM-instruct&2024 &4B &56.96  &69.64  &60.78  &60.53  &43.66  &62.75  &52.94 &52.00  &56.00  &54.00  &50.70  &59.26  &52.94  &56.32 \\
Deepseek-vl2&2024 & 4B & 53.16 &51.79 &64.71 &57.89
&59.15 &62.75 &52.94 &54.00 &46.00 &52.00 &60.56 &66.67 &42.00 &55.66 \\
Phi-3.5-vision&2024& 4.2B & 60.27 &75.00 &60.78 &64.00 &47.89 &68.63 &54.00 &56.00 &54.00 &54.00 &60.56 &64.81 &66.00 &60.45
   \\
LLaVA-v1.5-7B&2023  & 7B  & 49.37 &\textbf{76.79} &54.90 &46.67 &54.29
&56.00 &50.00 &52.00 &62.00 &46.00 &52.94 &50.94 &55.10 &54.38
   \\
LLaVA-v1.6-7B&2023 & 7B & 59.49 &69.64 &56.86 &65.79 &46.48
&62.75 &56.86 &50.00 &54.00 &60.00 &52.11 &61.11 &56.00 &57.77
   \\
LLaVA-OneVision&2024& 7B & 65.38 &66.07 &53.06 &57.33 &47.83
&58.82 &48.98 &50.00 &53.06 &54.00 &52.17 &53.70 &55.10 &55.03
   \\
Chameleon-7B&2024& 7B  & 52.11 &62.00 &65.96 &59.42 &52.38 &58.00 &45.83 &53.06 &55.32 &44.90 &67.19 &58.33 &55.56 &56.15
   \\
Qwen2-VL-7B-Instruct&2024 & 7B & 46.84 &55.36 &45.10 &47.37 &61.97 &50.98 &66.67 &56.00 &54.00 &50.00 &46.48 &57.41
&40.00  &52.16 \\
Qwen2.5-VL-7B&2025 & 7B   &60.76 &64.29 &60.78 &61.84 &60.56 &68.63 &70.59
&68.00 &64.00 &\textbf{72.00} &69.01 &68.52 &56.00 &64.99 \\
Molmo-7B&2024& 7B & 53.16 &50.00 &56.00 &60.00 &46.38 &45.10
&43.14 &58.00 &46.94 &50.00 &50.70 &56.60 &50.00 &51.23
   \\
MiniCPM-Llama3-V-2\_5&2024& 8B & 51.28 &51.79 &47.06 &50.67
&53.52 &45.10 &64.71 &60.00 &50.00 &58.00 &53.52
&64.15 &61.22 &54.69
   \\
MiniCPM-V2-6&2024& 8B & 49.09 &56.52 &56.10 &50.91 &52.08
&68.57 &62.50 &39.47 &51.22 &62.16 &60.38 &50.00 &59.09 &55.23
   \\
InternVL2-8B&2024& 8B  & 55.70 &58.93 &54.90 &50.00 &60.56
&47.06 &70.59 &46.00 &58.00 &60.00 &56.34 &57.41 &58.00 &56.42
   \\
InternVL2\_5-8B&2025  & 8B &63.29 &67.86 &50.98 &61.84 &53.52 &68.63
&60.78 &54.00 &38.00 &64.00 &57.75 &55.56 &52.00 &57.55 \\
InternVL3-8B &2025 &8B &48.10&62.50&\textbf{72.55}&67.11&56.34&66.67&66.67&\textbf{76.00}
&52.00&58.00&54.93&61.11&\textbf{68.00}&62.30 \\
GLM-4v&2024& 9B  & 62.03 &66.07 &49.02 &59.21 &61.97 &\textbf{74.51}
&68.63 &\textbf{76.00} &54.00 &64.00 &52.11 &62.96 &58.00&62.19
   \\
LLama-3.2-Vision&2024& 11B & 57.14   & 69.64 & 62.75  & 66.00 
& 57.14 & 41.07 & 64.71 & 54.90 & 50.00 & 56.00    & 40.00 & 51.79 & 62.96 & 54.00    \\
Pixtral-12B&2024& 12B & \textbf{65.82} &73.21 &47.06 &61.84 &56.34
&70.59 &60.78 &64.00 &62.00 &62.00 &54.93 &66.67 &56.00 &61.63
   \\
LLaVA-v1.5-13B& 2023& 13B & 58.33 &62.26 &57.14 &45.83 &56.92 &48.98 &48.98 &54.00 &47.83 &59.57 &56.72 &61.54
&51.06 &54.55 \\
InternVL-Chat-V1-5&2024 & 25.5B & 42.11 &51.79 &44.00 &51.32
&54.93 &56.86 &64.71 &60.00 &50.00 &64.00 &47.89 &64.81
&68.00 &55.41 \\
InternVL2-40B&2024 & 40B  & 62.03 &64.29 &62.75 &\textbf{68.42} &52.11
&70.59 &58.82 &50.00 &50.00 &60.00 &52.11 &61.11 &60.00 &59.40
   \\
Qwen2-VL-72B&2024& 72B &64.56 &62.50 &64.71 &65.79
&64.79 &60.78 &68.63 &72.00 &\textbf{66.00} &66.00
&60.56 &\textbf{72.22} &60.00 &65.27\\ \hline
\rowcolor{orange! 20}\multicolumn{17}{c}
{Novel Baselines} \\
Qwen2-VL-Judge&- &7B & 54.43 &66.07 &62.75 &61.84
&47.89 &52.94 &52.94 &62.00 &60.00 &58.00 &54.93 &59.26 &54.00 &57.46 \\ 
Qwen2-VL-DPO&- &7B &55.70 &53.57 &52.94 &55.26
&54.93 &47.06 &66.67 &52.00 &52.00 &60.00
&57.75 &50.00 &56.00 &54.91
 \\\hline
\rowcolor{green! 10}\multicolumn{17}{c}{Medical-Specific Models} \\
LLava-Med&2023 & 7B & 49.09&54.29&60.56&55.10&54.29
&60.78&50.00&50.00&62.00&62.00&47.89&50.94&55.10&54.77\\
STLLava-Med&2024 & 7B & 48.98 &54.29 &50.67 &61.97
&54.29 &60.78 &50.00 &50.00 &56.00 &62.00
&62.16 &50.94 &55.10 &55.16\\ 
HuatuoGPT-Vision&2024 & 7B & 59.74 &65.45 &54.90 &51.35
&60.61 &66.00 &58.00 &56.00 &46.81 &45.83 &52.86 &49.06 &56.00 &55.58 \\ 
Med-Flamingo&2023& 9B &46.94 &45.71 &49.33 &53.52
&45.71 &49.02 &60.00 &52.00 &56.00
&66.00 &48.65 &47.36 &47.14 &51.33 \\
MedDr& 2024& 40B & 59.49 &64.29 &60.78 &57.89 &46.48
&66.67 &60.78 &56.00  &54.00  &52.00  &50.70 
&62.96 &60.00 &57.84\\\hline
\rowcolor{pink! 20}\multicolumn{17}{c}{Proprietary MLLMs} \\
GPT-4o&-&/  & 48.10 &55.36 &64.71 &60.53 &56.34
&52.94 &49.02 &56.00 &56.00 &56.00 &61.97
&59.26 &56.00 &52.06 \\
Claude-3.5&- &/ & 53.16 &60.71 &62.75 &52.63 &60.56 &58.82 &58.82 &68.00 &60.00 &64.00 &59.15 &64.81 &60.00 &60.26 \\
Gemini-1.5-pro&-& /  & 55.37&61.89&68.63&\textbf{68.42}&62.45
&60.38&60.90&72.00 &61.00 &66.00&60.40&66.38&61.00 &63.51
  \\ 
O1&- & / &65.38 &73.21 &68.63 &67.19 &\textbf{68.42} &70.59 &\textbf{72.46} &70.00 &64.00 &66.00 &\textbf{71.50} &69.86 &\textbf{68.00} &\textbf{68.86}\\ \hline
\end{tabular}}
\caption{The overall performance of different MLLMs in judging, compared with human annotations on different organs. ABD, BRE, BRN, CHE, EYE, FOT, GI, HRT, LL, LNG, OC, PC, UL denote abdomen, brain, breast, chest, eye, foot, gastrointestinal tract, heart, lower limb, lung,  oral cavity, pelvic cavity, and upper limb, respectively.}
\label{tab:overall}
\end{table*}

\section{Experiments}

\subsection{Evaluated MLLMs}
We evaluated 32 state-of-the-art MLLMs, encompassing both open-source and proprietary models, as well as models specifically tailored for medical applications. The open-source models span a wide range of scales, from 3 billion to 72 billion parameters, and include: LLaVA (v1.5, v1.6, and LLaVA-OV)\cite{liu2024llava, li2024llava}, InternVL2 (8B, 40B, and Chat)\cite{chen2024far}, InternVL2\_5 (8B)~\cite{chen2024internvl}, InternVL3 (8B)~\cite{chen2024internvl}, Phi-3.5-Vision~\cite{abdin2024phi}, Qwen2-VL (7B and 72B)\cite{wang2024qwen2}, Qwen2.5-VL (7B)~\cite{Qwen2.5-VL}, LLaMA-3.2-Vision\cite{dubey2024llama}, Pixtral-12B~\cite{agrawal2024pixtral}, MiniCPM (v2.5 and v2.6)\cite{hu2024minicpm}, GLM-4V\cite{glm2024chatglm}, Chameleon (7B)\cite{Chameleon_Team_Chameleon_Mixed-Modal_Early-Fusion_2024}, Molmo-7B\cite{molmo2024}, VILA~\cite{lin2023vila}, xGen-MM~\cite{blip3}, and Deepseek-VL2-Small~\cite{wu2024deepseekvl2mixtureofexpertsvisionlanguagemodels}.
Among proprietary models, we included several widely recognized systems: GPT-4o~\cite{hurst2024gpt}, Claude-3.5-Sonnet, Gemini 1.5 Pro~\cite{team2024gemini}, and OpenAI O1.
In addition, we evaluated several models specifically designed for medical applications, including Med-Flamingo~\cite{alayrac2022flamingo}, LLaVA-Med~\cite{li2024llavamed}, STLLaVA-Med~\cite{sun2024self}, MedDr~\cite{he2024meddr}, and HuatuoGPT-Vision~\cite{chen2024huatuogpt}.

\subsection{Evaluation Settings}
We adopted a rigorous evaluation protocol based on the LLM-as-a-Judge paradigm. For each test sample, the model receives the same instruction that was provided to the human annotators.  The model’s ability to judge between the responses is assessed across six key dimensions.
All evaluations are conducted using fixed decoding parameters to ensure consistency across models.

\subsection{Empirical Results and Analysis}
\subsubsection{Results across Organs}
Table~\ref{tab:overall} presents a comprehensive evaluation of MLLM performance across diverse organ systems on~\ourmethod. The results reveal a clear hierarchical performance distribution among models. Proprietary systems and large-scale open-source MLLMs consistently outperform others, indicating a strong positive correlation between model size and overall capability.
O1 and Qwen2-VL-72B lead the top tier, achieving overall accuracy rates of 68.86\% and 65.27\%, respectively. The second tier includes Qwen2.5-VL-7B(64.99\%), Gemini (63.51\%), GLM-4V (62.19\%), and Pixtral-12B (61.63\%), all demonstrating competitive yet slightly lower performance.
Interestingly, despite being trained on specific medical datasets, medical-specific MLLMs exhibit comparatively limited judgment capabilities, with average accuracy rates of only 54–55\%. This suggests that current medical MLLMs, while proficient at recognizing standard medical features, struggle with judgment for different responses in complex reasoning.
Their performance gap likely reflects limited generalization and critical judgment abilities in non-routine settings, highlighting a pressing need for more robust training strategies and diversified data in the development of next-generation medical MLLMs.

The results also reveal substantial domain-specific variation in the judgment capabilities of different models across medical subfields. In cardiac imaging, Gemini and Qwen2-VL-72B achieved the highest accuracy (76\%), showing strong domain-specific reasoning. In gastrointestinal imaging, several models also performed well, with accuracy consistently exceeding 70\%, indicating their strength in systematic visual analysis and diagnostic precision.
However, performance dropped markedly in highly specialized domains. For example, in ophthalmology, even the top-performing models failed to surpass 70\% accuracy, underscoring current limitations of vision-language models in fields requiring exceptionally fine-grained judgment. These challenges likely arise from two main factors: (1) the depth and specificity of domain knowledge required, and (2) the inherent complexity of diagnostic criteria, which demand advanced reasoning and nuanced visual interpretation beyond the current capabilities of existing MLLMs.

\begin{figure}[t]
    \centering
    \includegraphics[width=0.9\columnwidth]{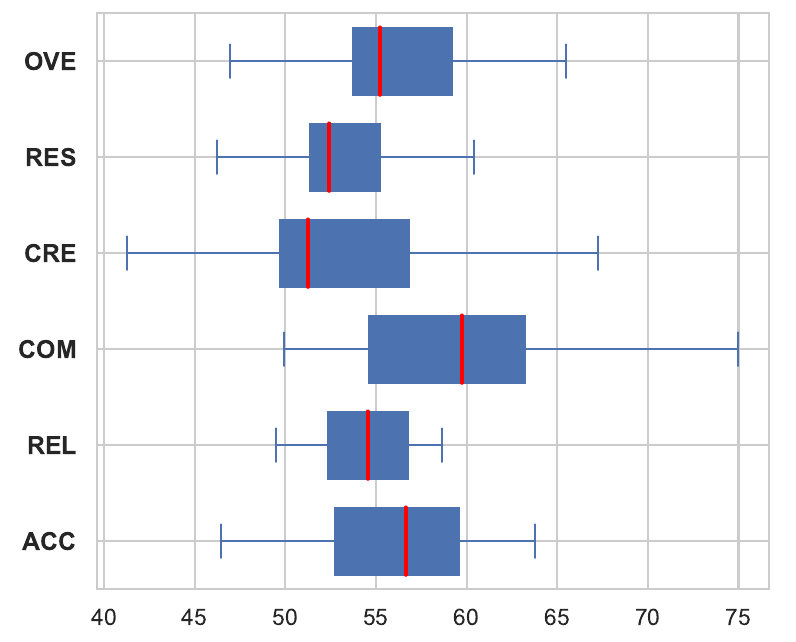}
    \caption{Box plots illustrate the distribution of performance scores for different MLLMs across six evaluation dimensions. The \textcolor{red}{red line} denotes median score, and the width of the boxes and whiskers shows the variability and spread of model performance across dimensions.}
    \label{fig:dimension}
\end{figure}

\begin{figure}[t]
\centering
\subfigure[Abdomen]{
\includegraphics[width=\columnwidth]{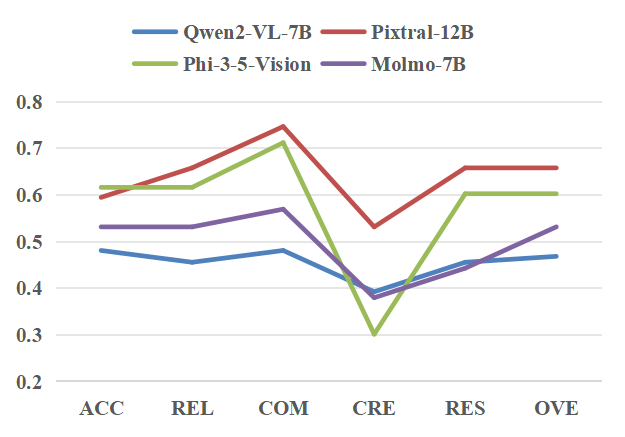}
\label{Fig2-1}
}\\
\subfigure[Brain]{
\includegraphics[width=\columnwidth]{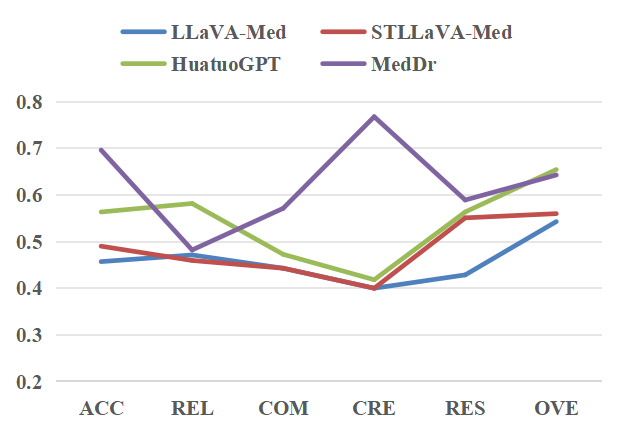}
\label{Fig2-4}
}
\caption{Performance of different MLLMs in abdomen (a) and rain (b) tasks across six evaluation dimensions.}
\label{Fig2}
\end{figure}

Fig.~\ref{fig:dimension} and Fig.~\ref{Fig2} illustrate the performance disparities among models across six evaluation dimensions. In Fig.~\ref{fig:dimension}, most models exhibit stable performance in accuracy (55\%) and responsiveness (50–60\%), reflecting a relatively mature capability in basic clinical judgment and timely response generation.
In contrast, substantial variation is observed in relevance, comprehensiveness, and creativity. For example, MedDr demonstrates marked deficiencies in relevance assessment for brain-related tasks in Fig.~\ref{Fig2} (b), while Phi-3.5-Vision performs poorly in creative interpretation for abdomen-related tasks in Fig.~\ref{Fig2} (a). For comprehensiveness, though most models approach 60\% in median score, high variance suggests inconsistent multi-dimensional reasoning capabilities across models in Fig.~\ref{fig:dimension}.
Overall performance scores range from 50\% to 65\%, indicating that current models remain at an intermediate level in synthesizing multi-dimensional judgments. These results underscore the ongoing challenges MLLMs face when addressing complex, unstructured clinical scenarios, and highlight significant opportunities for future improvement in reasoning, interpretability, and contextual understanding. Please see Appendix Tables~\ref{tab:abdomen} to~\ref{tab:Upper} for more results across different organs. 

\subsubsection{Results across Departments}
Table~\ref{tab:department} reveals substantial performance variability across medical departments among the evaluated models. Gastroenterology (GI) yields higher scores (O1: 75.95\%, Gemini-1.5: 74.68\%), suggesting these tasks benefit from MLLMs' strengths in contextual reasoning. Medical-specific models excel here, with MedDr achieving 71.23\%. Radiology (RAD) shows wide performance dispersion (46.24\%-65.59\%), indicating challenges in interpreting complex visual inputs. In general surgery (GS), Qwen2-VL-7B surprisingly leads (65.82\%), outperforming even proprietary models. Ophthalmology (OPH) results reveal sensitivity to fine-grained visual details, with scores ranging from 44.44\% (MiniCPM-V2-5) to 68.49\% (O1). Otolaryngology (ENT) demonstrates the most variability (41.03\%-68.29\%), suggesting unique multimodal reasoning challenges. Proprietary models perform consistently well, with O1 ranking first in three specialties and exceeding 62\% in all departments. Among open-source MLLMs, Qwen2-VL-72B shows the most balanced cross-specialty performance (53.66\%-64.56\%), while performance disparities across departments highlight the heterogeneous nature of medical visual reasoning tasks.

\begin{table}[ht]
\centering
\footnotesize
\renewcommand\arraystretch{1.2}
\tabcolsep=0.1cm
\scalebox{0.85}{
\begin{tabular}{l|lllll}
\hline
Model &OPH &RAD & ENT & GS  &GI   \\\hline
\rowcolor{gray! 10}\multicolumn{6}{c}{Open-Source MLLMs} \\
VILA1.5-3b &45.21 &58.57 &56.25 &52.17 &58.23   \\ 
xGen-MM-instruct &45.21 &59.14 &59.35 &53.26 &56.96   \\
Deepseek-vl2 &57.53 &52.97 &53.66 &54.13 &59.49 
 \\
Phi-3-5-Vision &49.32 &61.75 &56.10 &59.43 &60.27
 \\
LLaVA-1.5-7B &54.17 &53.79 &63.16 &46.79 &49.37
 \\
LLaVA-1.6-7B &46.58 &59.16 &48.78 &55.05 &59.49
 \\
LLaVA-OneVision-7B &47.14 &57.00 &46.34 &58.82
&65.38  \\
Chameleon-7B &53.85 &57.76 &56.41 &57.00 &52.11
 \\
Qwen2-VL-7B& 54.70 &51.22 &44.04 &\textbf{65.82} &57.64
\\
Qwen2.5-VL-7B &60.27 &60.64 &58.54 &57.80 &60.76  \\
Molmo-7B &47.89 &52.27 &42.50 &50.46 &53.16 \\
MiniCPM-V2-5 &44.44 &52.53 &42.11 &56.88 &54.43 
 \\
MiniCPM-V2-6 &53.73 &46.24 &52.63 &57.73 &47.22
 \\
InternVL2-8B &60.27 &56.93 &53.66 &55.96 &55.70
 \\
InternVL2\_5-8B &53.42 &55.69 &43.90 &52.29 &63.29  \\
InternVL3-8B &56.16 &61.63 &46.34 &56.88 &48.10 \\
GLM-4v &46.58 &55.97 &43.90 &44.44 &68.42
 \\
LLama-3.2-Vision &55.38 &49.43 &41.03
&48.00 &59.15 
 \\
Pixtral-12B &57.53 &61.14 &41.46 &55.96 &65.82
 \\
LLaVA-1.5-13B &58.21 &53.49 &47.37 &53.61 &58.33
 \\
InternVL-Chat-V1-5 &53.42 &51.74 &58.54 &54.63
&42.11  \\
InternVL2-40B & 53.42 &60.89 &43.90 &55.96 &62.03
 \\
Qwen2-VL-72B &64.38 &61.88 &53.66 &57.80 &64.56
 \\\hline
\rowcolor{green! 10}\multicolumn{6}{c}{Medical-Specific Models} \\
Med-Flamingo &49.30&56.31&32.50&50.46&53.16
 \\
LLava-Med &53.52 &53.54 &50.00 &62.39
&53.16 
 \\
STLLava-Med &53.42 &57.43 &56.10 &51.38
&67.09 
 \\
MedDr &56.16 &59.25 &63.41 &54.72
&71.23 \\
HuatuoGPT-Vision &52.05 &58.42 &46.34 &58.72
&58.23 
\\ \hline
\rowcolor{pink! 20}\multicolumn{6}{c}{Proprietary MLLMs} \\
GPT-4o &57.53 &55.71 &59.38 &57.61 &48.10  \\
Claude-1.5 &60.27 &56.86 &50.00 &59.78 &53.16  \\
Gemini-1.5 &60.27 &61.39 &\textbf{68.29} &57.80 &74.68 \\
O1 &\textbf{68.49} &\textbf{65.59} &58.54 &62.39 &\textbf{75.95}
\\\hline
\end{tabular}}
\caption{Overall performance of different MLLMs compared with human annotations across different departments. OPH, RAD, ENT, GS, and GI denote ophthalmology, radiology, otolaryngology, general surgery, and gastroenterology, respectively.}
\label{tab:department}
\end{table}

\subsubsection{Training Baselines}
To enhance the reward modeling capabilities of MLLMs, we fine-tune Qwen2-VL-7B using Supervised Fine-Tuning (SFT) and Direct Preference Optimization (DPO)~\cite{rafailov2023direct} on a curated training dataset of 10k image-question pairs, over 3 training epochs, within the LLaMA-Factory framework~\cite{zheng2024llamafactory}. We randomly select 10k difficult samples from the initial dataset $P$ in Section~\ref{step1_sec}, which are correctly answered by fewer than 3 MLLMs from 5 small MLLMs in Section~\ref{step1_sec}.
These samples do not overlap with~\ourmethod~ to avoid data leakage.
For the SFT stage (Qwen2-VL-Judge), we utilize high-quality responses generated by Qwen2-VL-72B as ground truth (GT) labels. In the DPO stage (Qwen2-VL-DPO), we construct preference pairs in which Qwen2-VL-72B responses serve as the "chosen" outputs, while responses from Qwen2-VL-2B serve as the "rejected" counterparts. To preserve the validity of preference learning, we exclude pairs with identical responses and ensure a balanced distribution of A/B label positions to mitigate positional bias. Table~\ref{tab:overall} shows Qwen2-VL-Judge and Qwen2-VL-DPO substantially improve the performance of the original Qwen2-VL-7B, indicating the effectiveness of the training process.

\section{Conclusion}
We introduce \ourmethod, the first comprehensive benchmark for evaluating reward models and judges in medical multimodal large language models. By integrating expert-annotated, clinically diverse cases and assessing six key dimensions of clinical quality, \ourmethod~reveals substantial challenges for both general and medical-specific MLLMs in aligning with expert judgment. Even current models show only moderate agreement with clinicians, underscoring the limitations of current approaches and the need for more rigorous evaluation standards. \ourmethod~and baseline results establish a foundation for advancing trustworthy and clinically aligned MLLMs.


\section*{Limitations}
In this study, we establish our baseline using the Qwen2-VL model, employing both Supervised Fine-Tuning (SFT) and Direct Preference Optimization (DPO) as training methodologies. Future research could explore alternative models as baselines and experiment with a wider range of training strategies to enhance model performance and broaden its applicability.

\bibliography{custom.bib}

\appendix
\clearpage
\textbf{\large Appendix for Med-RewardBench}

\vspace{10pt}
\noindent\textbf{Abstract.}
\label{sec:appendix}

Appendix~\ref{appendix:Prompt templates} presents the prompt templates used in Med-RewardBench.

Appendix~\ref{appendix:Physician Evaluation Guideline} outlines the physician evaluation guidelines applied during annotation.

Appendix~\ref{appendix:example} shows an example of Med-RewardBench in six dimensions.

Appendix~\ref{appendix:Results for Different Organs} details the results across 13 organs, evaluated along six distinct dimensions.

Appendix~\ref{appendix:Results for Different Departments} details the results across 8 departments, evaluated along six distinct dimensions.

\section{Prompt templates}
\label{appendix:Prompt templates}
\begin{figure*}[htbp]
    \centering
    \includegraphics[scale=0.7]{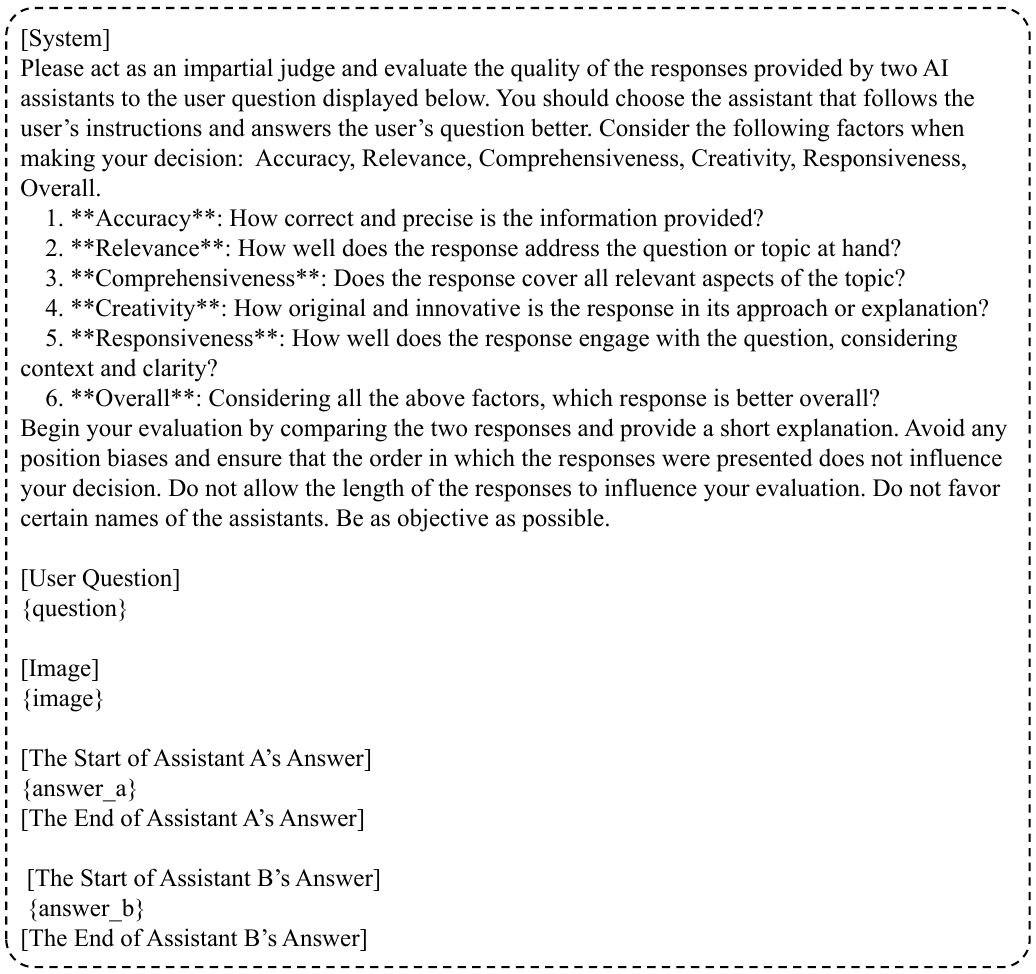}
    \caption{The default prompt for pairwise comparison.}
    \label{fig:prompt}
\end{figure*}

The prompt templates for \ourmethod{} consist of a system prompt, a user question, an image, and two candidate answers (A and B), as illustrated in Figure~\ref{fig:prompt}.

\section{Physician Evaluation Guideline}
\label{appendix:Physician Evaluation Guideline}
\begin{figure*}[htbp]
    \centering
    \includegraphics[scale=0.8]{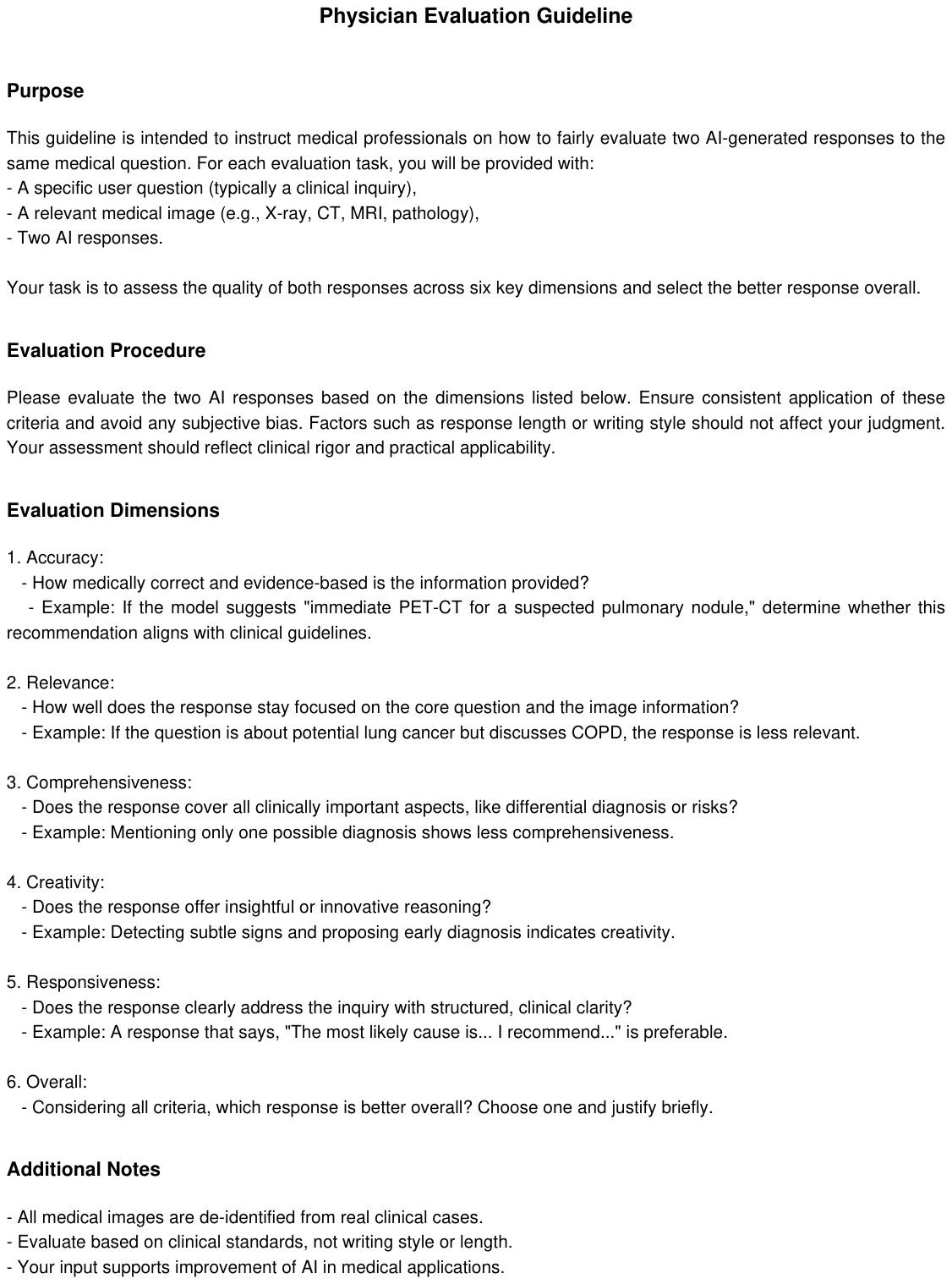}
    \caption{Annotation guidelines for medical experts.}
    \label{fig:guideline}
\end{figure*}

The physician evaluation guideline for \ourmethod~consist consists of purpose, evaluation procedure, evaluation dimensions, and additional notes, as illustrated in Figure~\ref{fig:guideline}.

\section{An example in \ourmethod}
\label{appendix:example}
\begin{figure*}[ht]
    \centering
    \includegraphics[scale=0.85]{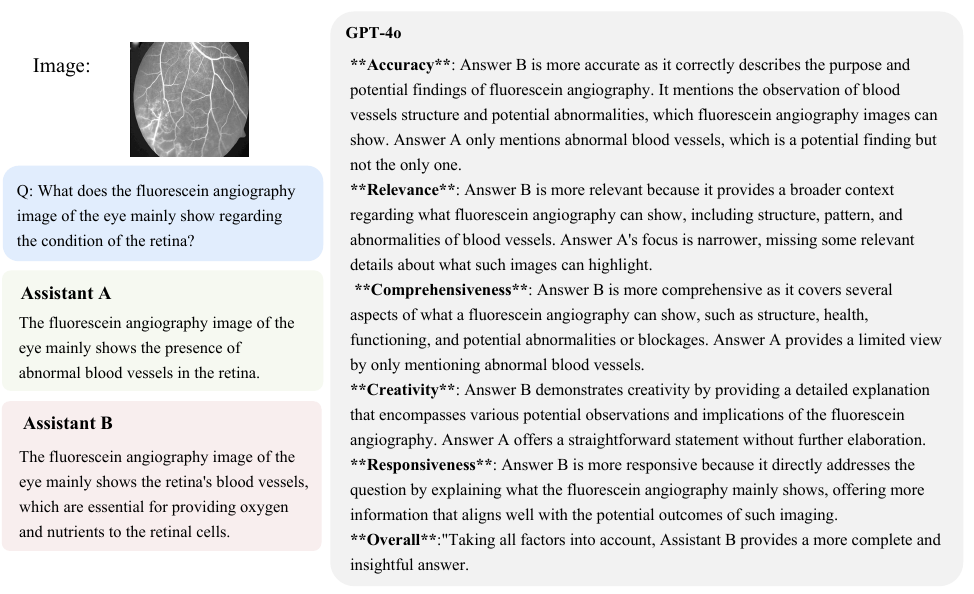}
    \caption{An example of \ourmethod~in six dimensions.}
    \label{fig:intro-example}
\end{figure*}

Fig~\ref{fig:intro-example} shows an example of \ourmethod~ in six dimensions.

\section{Results for Different Organs}
\label{appendix:Results for Different Organs}
Table~\ref{tab:abdomen} -~\ref{tab:Upper} show the details of each organs in six evaluation dimensions. ACC, REL, COM, CRE, RES, OVE denote accuracy, relevance, comprehensiveness, creativity, responsiveness, and overall, respectively.

\section{Results for Different Departments}
\label{appendix:Results for Different Departments}
Table~\ref{tab:department1} shows the details of each department in six evaluation dimensions.

\begin{table}[htbp]
\scalebox{0.6}{
\begin{tabular}{l|llllll}
\hline
Model & ACC & REL & COM & CRE & RES & OVE \\ \hline
\rowcolor{gray! 10}\multicolumn{7}{c}{Open-Source MLLMs} \\
VILA1.5-3b &56.96 	&62.03 	&63.29 	&62.03 	&62.03 	&58.23 \\ 
xGen-MM-instruct &62.03 &63.29 	&62.03 	&58.23 	&62.03 	&56.96 \\
Deepseek-vl2 & 53.16 & 58.23 & 50.63 & 35.44 & 51.90 & 53.16 \\
Phi-3-5-Vision & 61.64 & 61.64 & 71.23 & 30.14 & 60.27 & 60.27 \\
LLaVA-1.5-7B & 58.23 & 59.49 & 59.49 & 53.16 & 54.43 & 49.37 \\
LLaVA-1.6-7B & 68.35 & 62.03 & 67.09 & 40.51 & 63.29 & 59.49 \\
LLaVA-OneVision-7B & 56.41 & 61.54 & 69.23 & 48.72 & 52.56 & 65.38 \\
Chameleon-7B & 49.30 & 60.56 & 59.15 & 53.52 & 59.15 & 52.11 \\
Qwen2-VL-7B & 48.10 & 45.57 & 48.10 & 39.24 & 45.57 & 46.84 \\
Qwen2.5-VL-7B &63.29	&67.09	&72.15	&40.51	&54.43	&60.76 \\
Molmo-7B & 53.16 & 53.16 & 56.96 & 37.97 & 44.30 & 53.16 \\
MiniCPM-V2-5 & 50.00 & 53.85 & 58.97 & 43.59 & 50.00 & 51.28 \\
MiniCPM-V2-6 & 41.82 & 67.27 & 76.36 & 69.09 & 61.82 & 49.09 \\
InternVL2-8B & 58.23 & 70.89 & 59.49 & 60.76 & 50.63 & 55.70 \\
InternVL2\_5-8B &62.03	&69.62	&67.09	&56.96	&46.84	&63.29  \\
InternVL3-8B &55.70	&56.96	&70.89	&32.91	&58.23	&48.10 \\
GLM-4v & 63.29 & 60.76 & 81.01 & 36.71 & 56.96 & 62.03 \\
Pixtral-12B & 59.49 & 65.82 & 74.68 & 53.16 & 65.82 & 65.82 \\
LLaVA-1.5-13B & 40.28 & 61.11 & 59.72 & 58.33 & 47.22 & 58.33 \\
InternVL-Chat-V1-5 & 69.74 & 30.26 & 43.42 & 52.63 & 68.42 & 42.11 \\
InternVL2-40B & 58.23 & 68.35 & 74.68 & 74.68 & 55.70 & 62.03 \\
Qwen2-VL-72B & 65.82 & 65.82 & 75.95 & 69.62 & 65.82 & 64.56 \\ \hline
\rowcolor{orange! 10}\multicolumn{7}{c}{Novel Baselines} \\ 
Qwen2-VL-SFT & 55.70 & 55.70 & 64.56 & 63.29 & 60.76 & 54.43 \\
Qwen2-VL-DPO & 54.43 & 49.37 & 55.70 & 40.51 & 55.70 & 55.70 \\ \hline
\rowcolor{green! 10}\multicolumn{7}{c}{Medical-Specific Models} \\ 
Med-Flamingo & 48.98 & 42.86 & 55.10 & 42.86 & 42.86 & 46.94 \\
LLava-Med & 41.82 & 67.27 & 76.36 & 69.09 & 61.82 & 49.09 \\
STLLava-Med & 46.94 & 42.86 & 55.10 & 42.86 & 42.86 & 48.98 \\
MedDr & 56.96 & 65.82 & 67.09 & 68.35 & 60.76 & 59.49 \\
HuatuoGPT & 53.25 & 54.55 & 62.34 & 50.65 & 51.95 & 59.74 \\ \hline
\rowcolor{pink! 20}\multicolumn{7}{c}{Proprietary MLLMs} \\
GPT-4o & 59.49 & 59.49 & 73.41 & 68.35 & 62.03 & 48.10 \\
Claude-3.5 & 51.90 & 59.49 & 68.35 & 50.63 & 60.76 & 53.16 \\ 
Gemini-1.5 &54.23	&60.85	&69.12	&52.46	&62.01	&55.37\\
O1&64.85	&66.72	&69.14	&65.32	&67.89	&65.38
\\\hline
\end{tabular}}
\caption{Abdomen}
\label{tab:abdomen}
\end{table}

\begin{table}[htbp]
\scalebox{0.6}{
\begin{tabular}{l|llllll}
\hline
Model & ACC & REL & COM & CRE & RES & OVE \\ \hline
\rowcolor{gray! 10}\multicolumn{7}{c}{Open-Source MLLMs} \\
VILA1.5-3b &69.64 	&51.79 	&57.14 	&51.79 	&58.93 	&69.64 \\ 
xGen-MM-instruct &58.93 &57.14 	&51.79 	&69.64 	&51.79 	&69.64 \\ 
Deepseek-vl2 & 50.00 & 26.79 & 46.43 & 41.07 & 48.21 & 51.79 \\
Phi-3-5-Vision & 71.43 & 48.21 & 67.86 & 32.14 & 58.93 & 75.00 \\
LLaVA-1.5-7B & 62.50 & 55.36 & 58.93 & 41.07 & 48.21 & 76.79 \\
LLaVA-1.6-7B & 75.00 & 51.79 & 60.71 & 48.21 & 58.93 & 69.64 \\
LLaVA-OneVision-7B & 69.64 & 41.07 & 66.07 & 35.71 & 50.00 & 66.07 \\
Chameleon-7B & 28.00 & 44.00 & 60.00 & 54.00 & 58.00 & 62.00 \\
Qwen2-VL-7B & 57.14 & 46.43 & 62.50 & 44.64 & 55.36 & 55.36 \\
Qwen2.5-VL-7B &66.07	&46.43	&78.57	&42.86	&58.93	&64.29 \\
Molmo-7B & 57.14 & 51.79 & 55.36 & 44.64 & 48.21 & 50.00 \\
MiniCPM-V2-5 & 57.14 & 41.07 & 64.29 & 51.79 & 44.64 & 51.79 \\
MiniCPM-V2-6 & 63.04 & 45.65 & 71.74 & 56.52 & 65.22 & 56.52 \\
InternVL2-8B & 62.50 & 46.43 & 69.64 & 46.43 & 48.21 & 58.93 \\
InternVL2\_5-8B &64.29	&51.79	&64.29	&51.79	&41.07	&67.86  \\
InternVL3-8B &62.50	&55.36	&76.79	&46.43	&60.71	&62.50 \\
GLM-4v & 67.86 & 42.86 & 73.21 & 32.14 & 57.14 & 66.07 \\
Pixtral-12B & 73.21 & 51.79 & 69.64 & 60.71 & 60.71 & 73.21 \\
LLaVA-1.5-13B & 49.06 & 41.51 & 62.26 & 64.15 & 50.94 & 62.26 \\
InternVL-Chat-V1-5 & 46.43 & 53.57 & 64.29 & 55.36 & 50.00 & 51.79 \\
InternVL2-40B & 69.64 & 42.86 & 80.36 & 58.93 & 41.07 & 64.29 \\
Qwen2-VL-72B & 62.50 & 41.07 & 75.00 & 75.00 & 51.79 & 62.50 \\ \hline
\rowcolor{orange! 10}\multicolumn{7}{c}{Novel Baselines} \\ 
Qwen2-VL-SFT & 66.07 & 58.93 & 53.57 & 51.79 & 62.50 & 66.07 \\
Qwen2-VL-DPO & 53.57 & 44.64 & 64.29 & 48.21 & 55.36 & 53.57 \\ \hline
\rowcolor{green! 10}\multicolumn{7}{c}{Medical-Specific Models} \\ 
Med-Flamingo & 54.29 & 42.86 & 44.29 & 40.00 & 47.14 & 45.71 \\
LLava-Med & 45.71 & 47.14 & 44.29 & 40.00 & 42.86 & 54.29 \\
STLLava-Med & 45.71 & 47.14 & 44.29 & 40.00 & 42.86 & 54.29 \\
MedDr & 69.64 & 48.21 & 57.14 & 76.79 & 58.93 & 64.29 \\
HuatuoGPT & 56.36 & 58.18 & 47.27 & 41.82 & 56.36 & 65.45 \\ \hline
\rowcolor{pink! 20}\multicolumn{7}{c}{Proprietary MLLMs} \\
GPT-4o & 64.29 & 51.79 & 66.07 & 60.71 & 67.86 & 55.36 \\
Claude-3.5 & 58.93 & 57.14 & 51.79 & 51.79 & 60.71 & 60.71 \\
Gemini-1.5 &60.02	&58.67	&53.91	&53.28	&62.45	&61.89 \\
O1 &67.24	&68.51	&65.73	&66.39	&68.02	&73.21 \\
\hline
\end{tabular}}
\caption{Brain}
\label{tab:brain}
\end{table}

\begin{table}[htbp]
\scalebox{0.6}{
\begin{tabular}{l|llllll}
\hline
Model & ACC & REL & COM & CRE & RES & OVE \\ \hline
\rowcolor{gray! 10}\multicolumn{7}{c}{Open-Source MLLMs} \\
VILA1.5-3b &60.78 	&56.86 	&47.06 	&43.14 	&56.86 	&58.82 \\ 
xGen-MM-instruct  &56.86 &47.06 &56.86 	&58.82 	&43.14 	&60.78 \\ 
Deepseek-vl2 & 50.98 & 68.63 & 37.25 & 52.94 & 68.63 & 64.71 \\
Phi-3-5-Vision & 60.78 & 56.86 & 49.02 & 45.10 & 58.82 & 60.78 \\
LLaVA-1.5-7B & 52.94 & 49.02 & 33.33 & 43.14 & 52.94 & 54.90 \\
LLaVA-1.6-7B & 56.86 & 47.06 & 49.02 & 43.14 & 56.86 & 56.86 \\
LLaVA-OneVision-7B & 59.18 & 53.06 & 51.02 & 55.10 & 61.22 & 53.06 \\
Chameleon-7B & 48.94 & 46.81 & 53.19 & 46.81 & 36.17 & 65.96 \\
Qwen2-VL-7B & 49.02 & 47.06 & 50.98 & 58.82 & 41.18 & 45.10 \\
Qwen2.5-VL-7B &56.86	&62.75	&64.71	&56.86	&62.75	&60.78 \\
Molmo-7B & 52.00 & 60.00 & 58.00 & 54.00 & 54.00 & 56.00 \\
MiniCPM-V2-5 & 41.18 & 54.90 & 54.90 & 58.82 & 52.94 & 47.06 \\
MiniCPM-V2-6 & 56.10 & 43.90 & 56.10 & 53.66 & 58.54 & 56.10 \\
InternVL2-8B & 60.78 & 60.78 & 50.98 & 58.82 & 60.78 & 54.90 \\
InternVL2\_5-8B &66.67	&50.98	&49.02	&37.25	&50.98	&50.98 \\
InternVL3-8B &76.47	&70.59	&74.51	&29.41	&58.82	&72.55 \\
GLM-4v & 58.82 & 56.86 & 62.75 & 54.90 & 58.82 & 49.02 \\
Pixtral-12B & 54.90 & 54.90 & 50.98 & 64.71 & 43.14 & 47.06 \\
LLaVA-1.5-13B & 63.27 & 53.06 & 53.06 & 42.86 & 42.86 & 57.14 \\
InternVL-Chat-V1-5 & 54.00 & 44.00 & 50.00 & 54.00 & 62.00 & 44.00 \\
InternVL2-40B & 62.75 & 66.67 & 58.82 & 56.86 & 43.14 & 62.75 \\
Qwen2-VL-72B & 62.75 & 66.67 & 68.63 & 64.71 & 62.75 & 64.71 \\ \hline
\rowcolor{orange! 10}\multicolumn{7}{c}{Novel Baselines} \\ 
Qwen2-VL-SFT & 64.71 & 60.78 & 54.90 & 50.98 & 60.78 & 62.75 \\
Qwen2-VL-DPO & 45.10 & 52.94 & 56.86 & 56.86 & 47.06 & 52.94 \\ \hline
\rowcolor{green! 10}\multicolumn{7}{c}{Medical-Specific Models} \\ 
Med-Flamingo & 50.67 & 48.00 & 61.33 & 64.00 & 52.00 & 49.33 \\
LLava-Med & 43.66 & 47.89 & 67.61 & 53.52 & 56.34 & 60.56 \\
STLLava-Med & 49.33 & 52.00 & 61.33 & 64.00 & 48.00 & 50.67 \\
MedDr & 60.78 & 58.82 & 52.94 & 52.94 & 60.78 & 60.78 \\
HuatuoGPT & 50.98 & 50.98 & 47.06 & 49.02 & 47.06 & 54.90 \\ \hline
\rowcolor{pink! 20}\multicolumn{7}{c}{Proprietary MLLMs} \\
GPT-4o & 64.71 & 60.78 & 56.86 & 50.98 & 56.86 & 64.71 \\
Claude-3.5 & 62.75 & 64.71 & 52.94 & 52.94 & 58.82 & 62.75 \\
Gemini-1.5 &64.88	&66.25	&54.13	&55.06	&60.31	&68.63 \\
O1 &69.12	&70.00	&65.48	&66.79	&67.54	&68.63 \\
\hline
\end{tabular}}
\caption{Breast}
\label{tab:breast}
\end{table}

\begin{table}[htbp]
\scalebox{0.6}{
\begin{tabular}{l|llllll}
\hline
Model & ACC & REL & COM & CRE & RES & OVE \\ \hline
\rowcolor{gray! 10}\multicolumn{7}{c}{Open-Source MLLMs} \\
VILA1.5-3b &60.53 	&48.68 	&46.05 	&31.58 	&53.95 	&57.89 
 \\ 
xGen-MM-instruct  &53.95 	&46.05 	&48.68 	&57.89 	&31.58 	&60.53 
 \\
Deepseek-vl2 & 46.05 & 53.95 & 63.16 & 67.11 & 51.32 & 57.89 \\
Phi-3-5-Vision & 61.33 & 58.67 & 57.33 & 57.33 & 56.00 & 64.00 \\
LLaVA-1.5-7B & 65.33 & 49.33 & 58.67 & 42.67 & 46.67 & 46.67 \\
LLaVA-1.6-7B & 63.16 & 47.37 & 51.32 & 46.05 & 56.58 & 65.79 \\
LLaVA-OneVision-7B & 62.67 & 50.67 & 56.00 & 33.33 & 52.00 & 57.33 \\
Chameleon-7B & 49.28 & 47.83 & 56.52 & 42.03 & 50.72 & 59.42 \\
Qwen2-VL-7B & 47.37 & 53.95 & 57.89 & 68.42 & 52.63 & 47.37 \\
Qwen2.5-VL-7B &60.53	&56.58	&63.16	&55.26	&57.89	&61.84  \\
Molmo-7B & 60.00 & 50.67 & 54.67 & 62.67 & 48.00 & 60.00 \\
MiniCPM-V2-5 & 49.33 & 52.00 & 61.33 & 64.00 & 48.00 & 50.67 \\
MiniCPM-V2-6 & 61.82 & 45.45 & 56.36 & 29.09 & 43.64 & 50.91 \\
InternVL2-8B & 57.89 & 51.32 & 57.89 & 44.74 & 53.95 & 50.00 \\
InternVL2\_5-8B &61.84	&60.53	&57.89	&36.84	&53.95 &61.84 \\
InternVL3-8B &64.47	&63.16	&57.89	&48.68	&57.89	&67.11 \\
GLM-4v & 55.26 & 53.95 & 71.05 & 43.42 & 53.95 & 59.21 \\
Pixtral-12B & 52.63 & 48.68 & 57.89 & 47.37 & 60.53 & 61.84 \\
LLaVA-1.5-13B & 61.11 & 50.00 & 56.94 & 40.28 & 37.50 & 45.83 \\
InternVL-Chat-V1-5 & 46.05 & 56.58 & 68.42 & 57.89 & 43.42 & 51.32 \\
InternVL2-40B & 65.79 & 60.53 & 60.53 & 50.00 & 47.37 & 68.42 \\
Qwen2-VL-72B & 63.16 & 61.84 & 65.79 & 68.42 & 59.21 & 65.79 \\ \hline
\rowcolor{orange! 10}\multicolumn{7}{c}{Novel Baselines} \\ 
Qwen2-VL-SFT & 64.47 & 57.89 & 57.89 & 46.05 & 52.63 & 61.84 \\
Qwen2-VL-DPO & 55.26 & 51.32 & 67.11 & 68.42 & 47.37 & 55.26 \\ \hline
\rowcolor{green! 10}\multicolumn{7}{c}{Medical-Specific Models} \\ 
Med-Flamingo & 61.97 & 54.93 & 61.97 & 66.20 & 56.34 & 53.52 \\
LLava-Med & 57.14 & 53.06 & 46.94 & 51.02 & 48.98 & 55.10 \\
STLLava-Med & 53.52 & 56.34 & 61.97 & 66.20 & 54.93 & 61.97 \\
MedDr & 60.53 & 55.26 & 53.95 & 50.00 & 53.95 & 57.89 \\ \hline
\rowcolor{pink! 20}\multicolumn{7}{c}{Proprietary MLLMs} \\
HuatuoGPT & 47.30 & 52.70 & 51.35 & 54.05 & 51.35 & 51.35 \\
GPT-4o & 52.63 & 48.68 & 63.16 & 46.05 & 52.63 & 60.53 \\
Claude-3.5 & 51.32 & 56.58 & 64.47 & 44.74 & 52.63 & 52.63 \\
Gemini-1.5 &53.67	&58.2	&65.89	&47.13	&54.02	&68.42 \\
O1 &65.03	&67.42	&69.66	&64.55	&66.88	&67.19 \\
\hline
\end{tabular}}
\caption{Chest}
\label{tab:chest}
\end{table}

\begin{table}[htbp]
\scalebox{0.6}{
\begin{tabular}{l|llllll}
\hline
Model & ACC & REL & COM & CRE & RES & OVE \\ \hline
\rowcolor{gray! 10}\multicolumn{7}{c}{Open-Source MLLMs} \\
VILA1.5-3b &43.66 	&43.66 	&49.30 	&52.11 	&43.66 	&43.66 \\ 
xGen-MM-instruct  &43.66 	&49.30 	&43.66 	&43.66 	&52.11 	&43.66 \\
Deepseek-vl2 & 64.79 & 54.93 & 52.11 & 46.48 & 54.93 & 59.15 \\
Phi-3-5-Vision & 46.48 & 45.07 & 56.34 & 38.03 & 46.48 & 47.89 \\
LLaVA-1.5-7B & 45.71 & 47.14 & 44.29 & 40.00 & 42.86 & 54.29 \\
LLaVA-1.6-7B & 50.70 & 49.30 & 54.93 & 60.56 & 47.89 & 46.48 \\
LLaVA-OneVision-7B & 44.93 & 47.83 & 52.17 & 59.42 & 42.03 & 47.83 \\
Chameleon-7B & 53.97 & 55.56 & 49.21 & 60.32 & 55.56 & 52.38 \\
Qwen2-VL-7B & 63.38 & 66.20 & 54.93 & 47.89 & 60.56 & 61.97 \\
Qwen2.5-VL-7B &59.15	&59.15	&66.20	&47.89	&56.34	&60.56 \\
Molmo-7B & 52.17 & 49.28 & 49.28 & 56.52 & 42.03 & 46.38 \\
MiniCPM-V2-5 & 59.15 & 50.70 & 53.52 & 49.30 & 52.11 & 53.52 \\
MiniCPM-V2-6 & 54.17 & 47.92 & 56.25 & 58.33 & 43.75 & 52.08 \\
InternVL2-8B & 52.11 & 56.34 & 59.15 & 53.52 & 43.66 & 60.56 \\
InternVL2\_5-8B &49.30	&43.66	&53.52	&49.30	&45.07	&53.52 \\
InternVL3-8B &52.11	&52.11	&61.97	&45.07	&50.70	&56.34 \\
GLM-4v & 69.01 & 63.38 & 73.24 & 39.44 & 54.93 & 61.97 \\
Pixtral-12B & 53.52 & 52.11 & 66.20 & 57.75 & 60.56 & 56.34 \\
LLaVA-1.5-13B & 47.69 & 46.15 & 50.77 & 46.15 & 53.85 & 56.92 \\
InternVL-Chat-V1-5 & 53.52 & 52.11 & 56.34 & 40.85 & 46.48 & 54.93 \\
InternVL2-40B & 52.11 & 46.48 & 61.97 & 50.70 & 46.48 & 52.11 \\
Qwen2-VL-72B & 64.79 & 61.97 & 70.42 & 61.97 & 64.79 & 64.79 \\ \hline
\rowcolor{orange! 10}\multicolumn{7}{c}{Novel Baselines} \\ 
Qwen2-VL-SFT & 47.89 & 50.70 & 53.52 & 56.34 & 47.89 & 47.89 \\
Qwen2-VL-DPO & 57.75 & 64.79 & 60.56 & 47.89 & 54.93 & 54.93 \\ \hline
\rowcolor{green! 10}\multicolumn{7}{c}{Medical-Specific Models} \\ 
Med-Flamingo & 54.29 & 42.86 & 44.29 & 40.00 & 47.14 & 45.71 \\
LLava-Med & 45.71 & 47.14 & 44.29 & 40.00 & 42.86 & 54.29 \\
STLLava-Med & 45.71 & 47.14 & 44.29 & 40.00 & 42.86 & 54.29 \\
MedDr & 43.66 & 52.11 & 50.70 & 56.34 & 45.07 & 46.48 \\
HuatuoGPT & 51.52 & 43.94 & 57.58 & 45.45 & 53.03 & 60.61 \\ \hline
\rowcolor{pink! 20}\multicolumn{7}{c}{Proprietary MLLMs} \\
GPT-4o & 50.70 & 59.15 & 56.34 & 50.70 & 67.61 & 56.34 \\
Claude-3.5 & 43.66 & 47.89 & 67.61 & 53.52 & 56.34 & 60.56 \\
Gemini-1.5 &46.51	&50.12	&68.90 	&55.76	&58.70 	&62.45 \\
O1 &63.72	&65.9	&69.82	&66.23	&67.41	&68.42 \\
\hline
\end{tabular}}
\caption{Eye}
\label{tab:eye}
\end{table}

\begin{table}[htbp]
\scalebox{0.6}{
\begin{tabular}{l|llllll}
\hline
Model & ACC & REL & COM & CRE & RES & OVE \\ \hline
\rowcolor{gray! 10}\multicolumn{7}{c}{Open-Source MLLMs} \\
VILA1.5-3b &62.75 	&60.78 	&58.82 	&49.02 	&58.82 	&62.75 \\ 
xGen-MM-instruct  &58.82 	&58.82 	&60.78 	&62.75 	&49.02 	&62.75 \\
Deepseek-vl2 & 45.10 & 64.71 & 58.82 & 41.18 & 56.86 & 62.75 \\
Phi-3-5-Vision & 60.78 & 68.63 & 64.71 & 41.18 & 60.78 & 68.63 \\
LLaVA-1.5-7B & 54.00 & 60.00 & 58.00 & 38.00 & 60.00 & 56.00 \\
LLaVA-1.6-7B & 60.78 & 62.75 & 60.78 & 45.10 & 60.78 & 62.75 \\
LLaVA-OneVision-7B & 58.82 & 54.90 & 60.78 & 37.25 & 54.90 & 58.82 \\
Chameleon-7B & 36.00 & 64.00 & 56.00 & 56.00 & 58.00 & 58.00 \\
Qwen2-VL-7B & 47.06 & 54.90 & 54.90 & 50.98 & 45.10 & 50.98 \\
Qwen2.5-VL-7B &60.78	&72.55	&64.71	&49.02	&66.67	&68.63 \\
Molmo-7B & 62.75 & 43.14 & 58.82 & 60.78 & 58.82 & 45.10 \\
MiniCPM-V2-5 & 35.29 & 49.02 & 52.94 & 52.94 & 41.18 & 45.10 \\
MiniCPM-V2-6 & 31.43 & 65.71 & 65.71 & 45.71 & 42.86 & 68.57 \\
InternVL2-8B & 50.98 & 58.82 & 50.98 & 58.82 & 50.98 & 47.06 \\
InternVL2\_5-8B &62.75	&60.78	&66.67	&50.98	&49.02	&68.63 \\
InternVL3-8B &49.02	&60.78	&68.63	&47.06	&58.82	&66.67 \\
GLM-4v & 68.63 & 72.55 & 70.59 & 29.41 & 54.90 & 74.51 \\
Pixtral-12B & 64.71 & 76.47 & 66.67 & 56.86 & 56.86 & 70.59 \\
LLaVA-1.5-13B & 53.06 & 63.27 & 57.14 & 53.06 & 46.94 & 48.98 \\
InternVL-Chat-V1-5 & 49.02 & 54.90 & 52.94 & 50.98 & 50.98 & 56.86 \\
InternVL2-40B & 64.71 & 70.59 & 72.55 & 66.67 & 50.98 & 70.59 \\
Qwen2-VL-72B & 49.02 & 66.67 & 64.71 & 66.67 & 60.78 & 60.78 \\ \hline
\rowcolor{orange! 10}\multicolumn{7}{c}{Novel Baselines} \\ 
Qwen2-VL-SFT & 49.02 & 58.82 & 49.02 & 58.82 & 49.02 & 52.94 \\
Qwen2-VL-DPO & 45.10 & 50.98 & 50.98 & 52.94 & 45.10 & 47.06 \\ \hline
\rowcolor{green! 10}\multicolumn{7}{c}{Medical-Specific Models} \\ 
Med-Flamingo & 60.78 & 60.78 & 64.71 & 66.67 & 66.67 & 49.02 \\
LLava-Med & 49.02 & 66.67 & 64.71 & 66.67 & 60.78 & 60.78 \\
STLLava-Med & 49.02 & 66.67 & 64.71 & 66.67 & 60.78 & 60.78 \\
MedDr & 62.75 & 76.47 & 60.78 & 64.71 & 60.78 & 66.67 \\
HuatuoGPT & 62.00 & 50.00 & 58.00 & 46.00 & 54.00 & 66.00 \\ \hline
\rowcolor{pink! 20}\multicolumn{7}{c}{Proprietary MLLMs} \\
GPT-4o & 52.94 & 62.75 & 54.90 & 62.75 & 58.82 & 52.94 \\
Claude-3.5 & 58.82 & 76.47 & 58.82 & 52.94 & 68.63 & 58.82 \\
Gemini-1.5 &60.70 	&78.03	&60.51	&55.42	&70.05	&60.38 \\
O1 &68.14	&69.88	&67.41	&66.90 	&69.27	&70.59 \\
\hline
\end{tabular}}
\caption{Foot}
\label{tab:foot}
\end{table}

\begin{table}[htbp]
\scalebox{0.6}{
\begin{tabular}{l|llllll}
\hline
Model & ACC & REL & COM & CRE & RES & OVE \\ \hline
\rowcolor{gray! 10}\multicolumn{7}{c}{Open-Source MLLMs} \\
VILA1.5-3b &52.94 	&43.14 	&56.86 	&62.75 	&49.02 	&52.94 \\ 
xGen-MM-instruct  &57.41 	&51.85 	&53.70 	&59.26 	&50.00 	&59.26 \\
Deepseek-vl2 & 64.71 & 50.98 & 58.82 & 33.33 & 49.02 & 52.94 \\
Phi-3-5-Vision & 52.00 & 42.00 & 60.00 & 34.00 & 52.00 & 54.00 \\
LLaVA-1.5-7B & 60.00 & 48.00 & 50.00 & 52.00 & 54.00 & 50.00 \\
LLaVA-1.6-7B & 66.67 & 50.98 & 58.82 & 39.22 & 49.02 & 56.86 \\
LLaVA-OneVision-7B & 46.94 & 42.86 & 55.10 & 42.86 & 42.86 & 48.98 \\
Chameleon-7B & 60.42 & 54.17 & 64.58 & 62.50 & 50.00 & 45.83 \\
Qwen2-VL-7B & 64.71 & 47.06 & 60.78 & 39.22 & 41.18 & 66.67 \\
Qwen2.5-VL-7B &70.59	&49.02	&82.35	&37.25	&39.22	&70.59 \\
Molmo-7B & 66.67 & 49.02 & 68.63 & 45.10 & 47.06 & 43.14 \\
MiniCPM-V2-5 & 64.71 & 60.78 & 70.59 & 49.02 & 50.98 & 64.71 \\
MiniCPM-V2-6 & 52.50 & 40.00 & 67.50 & 55.00 & 40.00 & 62.50 \\
InternVL2-8B & 58.82 & 52.94 & 62.75 & 70.59 & 58.82 & 70.59 \\
InternVL2\_5-8B &54.90	&47.06	&54.90	&52.94	&62.75	&60.78 \\
InternVL3-8B &72.55	&43.14	&80.39	&45.10	&49.02	&66.67 \\
GLM-4v & 64.71 & 49.02 & 76.47 & 47.06 & 50.98 & 68.63 \\
Pixtral-12B & 70.59 & 41.18 & 68.63 & 68.63 & 52.94 & 60.78 \\
LLaVA-1.5-13B & 46.94 & 38.78 & 65.31 & 63.27 & 46.94 & 48.98 \\
InternVL-Chat-V1-5 & 66.67 & 56.86 & 54.90 & 58.82 & 52.94 & 64.71 \\
InternVL2-40B & 58.82 & 45.10 & 82.35 & 80.39 & 58.82 & 58.82 \\
Qwen2-VL-72B & 68.63 & 50.98 & 78.43 & 68.63 & 45.10 & 68.63 \\ \hline
\rowcolor{orange! 10}\multicolumn{7}{c}{Novel Baselines} \\ 
Qwen2-VL-SFT & 52.94 & 50.98 & 52.94 & 50.98 & 49.02 & 52.94 \\
Qwen2-VL-DPO & 60.78 & 43.14 & 68.63 & 37.25 & 41.18 & 66.67 \\ \hline
\rowcolor{green! 10}\multicolumn{7}{c}{Medical-Specific Models} \\ 
Med-Flamingo & 50.00 & 54.00 & 50.00 & 52.00 & 48.00 & 60.00 \\
LLava-Med & 60.00 & 48.00 & 50.00 & 52.00 & 54.00 & 50.00 \\
STLLava-Med & 60.00 & 48.00 & 50.00 & 52.00 & 54.00 & 50.00 \\
MedDr & 52.94 & 56.86 & 66.67 & 60.78 & 45.10 & 60.78 \\
HuatuoGPT & 54.00 & 40.00 & 52.00 & 52.00 & 52.00 & 58.00 \\ \hline
\rowcolor{pink! 20}\multicolumn{7}{c}{Proprietary MLLMs} \\
GPT-4o & 56.86 & 52.94 & 74.51 & 62.75 & 39.22 & 49.02 \\
Claude-3.5 & 52.94 & 56.86 & 56.86 & 52.94 & 43.14 & 58.82 \\
Gemini-1.5 &55.19	&58.72	&59.13	&54.80 	&45.23	&60.90 \\
O1 &66.81	&68.02	&67.23	&65.79	&64.96	&72.46 \\
\hline
\end{tabular}}
\caption{Gastrointestinal tract}
\label{tab:gastro}
\end{table}

\begin{table}[htbp]
\scalebox{0.6}{
\begin{tabular}{l|llllll}
\hline
Model & ACC & REL & COM & CRE & RES & OVE \\ \hline
\rowcolor{gray! 10}\multicolumn{7}{c}{Open-Source MLLMs} \\
VILA1.5-3b &52.00 	&58.00 	&52.00 	&46.00 	&56.00 	&52.00 \\ 
xGen-MM-instruct  &56.00 	&52.00 	&58.00 	&52.00 	&46.00 	&52.00 \\
Deepseek-vl2 & 52.00 & 50.00 & 66.00 & 54.00 & 52.00 & 54.00 \\
Phi-3-5-Vision & 56.00 & 42.00 & 58.00 & 36.00 & 58.00 & 56.00 \\
LLaVA-1.5-7B & 54.00 & 54.00 & 38.00 & 52.00 & 64.00 & 52.00 \\
LLaVA-1.6-7B & 52.00 & 52.00 & 54.00 & 44.00 & 58.00 & 50.00 \\
LLaVA-OneVision-7B & 54.00 & 64.00 & 58.00 & 48.00 & 56.00 & 50.00 \\
Chameleon-7B & 59.18 & 42.86 & 51.02 & 57.14 & 48.98 & 53.06 \\
Qwen2-VL-7B & 56.00 & 42.00 & 52.00 & 54.00 & 56.00 & 56.00 \\
Qwen2.5-VL-7B &68.00	&48.00	&70.00	&46.00	&52.00	&68.00 \\
Molmo-7B & 56.00 & 64.00 & 48.00 & 52.00 & 54.00 & 58.00 \\
MiniCPM-V2-5 & 58.00 & 38.00 & 68.00 & 56.00 & 58.00 & 60.00 \\
MiniCPM-V2-6 & 50.00 & 55.26 & 63.16 & 50.00 & 55.26 & 39.47 \\
InternVL2-8B & 50.00 & 48.00 & 66.00 & 48.00 & 66.00 & 46.00 \\
InternVL2\_5-8B &56.00	&62.00	&60.00	&44.00	&50.00	&54.00 \\
InternVL3-8B &74.00	&62.00	&76.00	&34.00	&56.00	&76.00 \\
GLM-4v & 66.00 & 44.00 & 70.00 & 34.00 & 58.00 & 76.00 \\
Pixtral-12B & 56.00 & 56.00 & 58.00 & 60.00 & 54.00 & 64.00 \\
LLaVA-1.5-13B & 50.00 & 54.00 & 58.00 & 54.00 & 44.00 & 54.00 \\
InternVL-Chat-V1-5 & 60.00 & 54.00 & 60.00 & 52.00 & 50.00 & 60.00 \\
InternVL2-40B & 54.00 & 56.00 & 68.00 & 62.00 & 58.00 & 50.00 \\
Qwen2-VL-72B & 72.00 & 42.00 & 86.00 & 74.00 & 56.00 & 72.00 \\ \hline
\rowcolor{orange! 10}\multicolumn{7}{c}{Novel Baselines} \\ 
Qwen2-VL-SFT & 62.00 & 52.00 & 54.00 & 56.00 & 62.00 & 62.00 \\
Qwen2-VL-DPO & 50.00 & 48.00 & 50.00 & 56.00 & 50.00 & 52.00 \\ \hline
\rowcolor{green! 10}\multicolumn{7}{c}{Medical-Specific Models} \\ 
Med-Flamingo & 50.00 & 58.00 & 54.00 & 44.00 & 52.00 & 52.00 \\
LLava-Med & 52.00 & 52.00 & 54.00 & 44.00 & 58.00 & 50.00 \\
STLLava-Med & 52.00 & 52.00 & 54.00 & 44.00 & 58.00 & 50.00 \\
MedDr & 52.00 & 46.00 & 56.00 & 66.00 & 56.00 & 5600.00 \\
HuatuoGPT & 66.00 & 44.00 & 62.00 & 44.00 & 56.00 & 56.00 \\ \hline
\rowcolor{pink! 20}\multicolumn{7}{c}{Proprietary MLLMs} \\
GPT-4o & 50.00 & 58.00 & 68.00 & 56.00 & 46.00 & 56.00 \\
Claude-3.5 & 52.00 & 58.00 & 66.00 & 56.00 & 50.00 & 68.00 \\
Gemini-1.5 &54.00 	&60.00 	&67.00 	&58.00 	&52.00 	&72.00 \\
O1 &66.00 	&67.00 	&70.00 	&68.00 	&65.00 	&70.00 \\ 
\hline
\end{tabular}}
\caption{Heart}
\label{tab:heart}
\end{table}

\begin{table}[htbp]
\scalebox{0.6}{
\begin{tabular}{l|llllll}
\hline
Model & ACC & REL & COM & CRE & RES & OVE \\ \hline
\rowcolor{gray! 10}\multicolumn{7}{c}{Open-Source MLLMs} \\
VILA1.5-3b &56.00 	&46.00 	&60.00 	&54.00 	&50.00 	&52.00 \\ 
xGen-MM-instruct  &52.00 	&42.00 	&58.00 	&52.00 	&46.00 	&54.00 \\
Deepseek-vl2 & 52.00 & 54.00 & 52.00 & 50.00 & 44.00 & 46.00 \\
Phi-3-5-Vision & 58.00 & 58.00 & 58.00 & 42.00 & 50.00 & 54.00 \\
LLaVA-1.5-7B & 48.00 & 46.00 & 42.00 & 68.00 & 46.00 & 62.00 \\
LLaVA-1.6-7B & 56.00 & 56.00 & 62.00 & 54.00 & 52.00 & 54.00 \\
LLaVA-OneVision-7B & 59.18 & 44.90 & 61.22 & 40.82 & 44.90 & 53.06 \\
Chameleon-7B & 38.30 & 53.19 & 46.81 & 51.06 & 55.32 & 55.32 \\
Qwen2-VL-7B & 48.00 & 54.00 & 54.00 & 50.00 & 52.00 & 54.00 \\
Qwen2.5-VL-7B &64.00	&58.00	&66.00	&40.00	&56.00	&64.0 \\
Molmo-7B & 57.14 & 46.94 & 53.06 & 55.10 & 44.90 & 46.94 \\
MiniCPM-V2-5 & 48.00 & 54.00 & 50.00 & 44.00 & 48.00 & 50.00 \\
MiniCPM-V2-6 & 68.29 & 48.78 & 73.17 & 60.98 & 43.90 & 51.22 \\
InternVL2-8B & 62.00 & 44.00 & 46.00 & 44.00 & 48.00 & 58.00 \\
InternVL2\_5-8B &52.00	&46.00	&58.00	&52.00	&46.00	&38.00 \\
InternVL3-8B &62.00	&50.00	&62.00	&46.00	&50.00	&52.00 \\
GLM-4v & 64.00 & 56.00 & 60.00 & 40.00 & 48.00 & 54.00 \\
Pixtral-12B & 62.00 & 52.00 & 68.00 & 58.00 & 58.00 & 62.00 \\
LLaVA-1.5-13B & 58.70 & 47.83 & 50.00 & 60.87 & 47.83 & 47.83 \\
InternVL-Chat-V1-5 & 58.00 & 56.00 & 58.00 & 40.00 & 44.00 & 50.00 \\
InternVL2-40B & 52.00 & 58.00 & 68.00 & 50.00 & 54.00 & 50.00 \\
Qwen2-VL-72B & 66.00 & 60.00 & 68.00 & 68.00 & 64.00 & 66.00 \\ \hline
\rowcolor{orange! 10}\multicolumn{7}{c}{Novel Baselines} \\ 
Qwen2-VL-SFT & 60.00 & 54.00 & 64.00 & 62.00 & 58.00 & 60.00 \\
Qwen2-VL-DPO & 48.00 & 54.00 & 58.00 & 48.00 & 52.00 & 52.00 \\ \hline
\rowcolor{green! 10}\multicolumn{7}{c}{Medical-Specific Models} \\ 
Med-Flamingo & 56.00 & 56.00 & 52.00 & 54.00 & 42.00 & 56.00 \\
LLava-Med & 48.00 & 46.00 & 42.00 & 68.00 & 46.00 & 62.00 \\
STLLava-Med & 56.00 & 42.00 & 52.00 & 54.00 & 56.00 & 56.00 \\
MedDr & 58.00 & 54.00 & 62.00 & 74.00 & 52.00 & 54.00 \\
HuatuoGPT & 53.19 & 44.68 & 57.45 & 55.32 & 53.19 & 46.81 \\ \hline
\rowcolor{pink! 20}\multicolumn{7}{c}{Proprietary MLLMs} \\
GPT-4o & 52.00 & 54.00 & 72.00 & 62.00 & 38.00 & 56.00 \\
Claude-3.5 & 56.00 & 48.00 & 62.00 & 54.00 & 52.00 & 60.00 \\
Gemini-1.5 &58.00 	&50.00 	&64.00 	&56.00 	&54.00 	&61.00 \\
O1 &67.00 	&66.00 	&69.00 	&66.00 	&67.00 	&64.00 \\ 
\hline
\end{tabular}}
\caption{Lower\_limb}
\label{tab:lowerlimb}
\end{table}

\begin{table}[htbp]
\scalebox{0.6}{
\begin{tabular}{l|llllll}
\hline
Model & ACC & REL & COM & CRE & RES & OVE \\ \hline
\rowcolor{gray! 10}\multicolumn{7}{c}{Open-Source MLLMs} \\
VILA1.5-3b &54.00 	&58.00 	&42.00 	&46.00 	&52.00 	&52.00 \\ 
xGen-MM-instruct  &50.00 	&60.00 	&46.00 	&52.00 	&54.00 	&56.00 \\
Deepseek-vl2 & 56.00 & 42.00 & 62.00 & 50.00 & 46.00 & 52.00 \\
Phi-3-5-Vision & 54.00 & 58.00 & 48.00 & 60.00 & 52.00 & 54.00 \\
LLaVA-1.5-7B & 44.00 & 46.00 & 42.00 & 58.00 & 54.00 & 46.00 \\
LLaVA-1.6-7B & 56.00 & 54.00 & 64.00 & 48.00 & 54.00 & 60.00 \\
LLaVA-OneVision-7B & 56.00 & 66.00 & 42.00 & 38.00 & 48.00 & 54.00 \\
Chameleon-7B & 46.94 & 38.78 & 55.10 & 59.18 & 57.14 & 44.90 \\
Qwen2-VL-7B & 54.00 & 62.00 & 58.00 & 54.00 & 42.00 & 50.00 \\
Qwen2.5-VL-7B &68.00	&64.00	&74.00	&42.00	&46.00	&72.00 \\
Molmo-7B & 56.00 & 50.00 & 58.00 & 58.00 & 54.00 & 50.00 \\
MiniCPM-V2-5 & 58.00 & 42.00 & 68.00 & 48.00 & 58.00 & 58.00 \\
MiniCPM-V2-6 & 48.65 & 45.95 & 43.24 & 43.24 & 45.95 & 62.16 \\
InternVL2-8B & 62.00 & 58.00 & 60.00 & 54.00 & 56.00 & 60.00 \\
InternVL2\_5-8B &64.00	&54.00	&50.00	&52.00	&48.00	&64.00 \\
InternVL3-8B &62.00	&62.00	&68.00	&44.00	&56.00	&58.00 \\
GLM-4v & 66.00 & 64.00 & 68.00 & 64.00 & 50.00 & 64.00 \\
Pixtral-12B & 66.00 & 60.00 & 66.00 & 56.00 & 54.00 & 62.00 \\
LLaVA-1.5-13B & 53.19 & 57.45 & 48.94 & 55.32 & 53.19 & 59.57 \\
InternVL-Chat-V1-5 & 64.00 & 50.00 & 50.00 & 44.00 & 62.00 & 64.00 \\
InternVL2-40B & 48.00 & 58.00 & 66.00 & 60.00 & 58.00 & 60.00 \\
Qwen2-VL-72B & 64.00 & 44.00 & 84.00 & 66.00 & 54.00 & 66.00 \\ \hline
\rowcolor{orange! 10}\multicolumn{7}{c}{Novel Baselines} \\ 
Qwen2-VL-SFT & 60.00 & 56.00 & 52.00 & 56.00 & 54.00 & 58.00 \\
Qwen2-VL-DPO & 62.00 & 52.00 & 66.00 & 56.00 & 60.00 & 60.00 \\ \hline
\rowcolor{green! 10}\multicolumn{7}{c}{Medical-Specific Models} \\ 
Med-Flamingo & 62.00 & 54.00 & 66.00 & 56.00 & 60.00 & 66.00 \\
LLava-Med & 48.00 & 46.00 & 42.00 & 68.00 & 46.00 & 62.00 \\
STLLava-Med & 66.00 & 60.00 & 66.00 & 56.00 & 54.00 & 62.00 \\
MedDr & 54.00 & 50.00 & 50.00 & 48.00 & 54.00 & 52.00 \\
HuatuoGPT & 41.67 & 56.25 & 41.67 & 54.17 & 43.75 & 45.83 \\ \hline
\rowcolor{pink! 20}\multicolumn{7}{c}{Proprietary MLLMs} \\
GPT-4o & 64.00 & 62.00 & 62.00 & 52.00 & 62.00 & 56.00 \\
Claude-3.5 & 62.00 & 52.00 & 54.00 & 54.00 & 54.00 & 64.00 \\
Gemini-1.5 &63.00 	&54.00 	&56.00 	&56.00 	&57.00 	&66.00 \\
O1 &68.00 	&66.00 	&67.00 	&67.00 	&67.00 	&66.00 \\ 
\hline
\end{tabular}}
\caption{Lung}
\label{tab:lung}
\end{table}

\begin{table}[htbp]
\scalebox{0.6}{
\begin{tabular}{l|llllll}
\hline
Model & ACC & REL & COM & CRE & RES & OVE \\ \hline
\rowcolor{gray! 10}\multicolumn{7}{c}{Open-Source MLLMs} \\
VILA1.5-3b &50.70 	&52.11 	&49.30 	&50.70 	&49.30 	&52.11 \\ 
xGen-MM-instruct  &49.30 	&49.30 	&52.11 	&52.11 	&50.70 	&50.70 \\
Deepseek-vl2 & 54.93 & 53.52 & 61.97 & 42.25 & 56.34 & 60.56 \\
Phi-3-5-Vision & 59.15 & 63.38 & 56.34 & 45.07 & 52.11 & 60.56 \\
LLaVA-1.5-7B & 51.47 & 52.94 & 50.00 & 50.00 & 51.47 & 52.94 \\
LLaVA-1.6-7B & 53.52 & 50.70 & 50.70 & 43.66 & 50.70 & 52.11 \\
LLaVA-OneVision-7B & 55.07 & 52.17 & 55.07 & 53.62 & 60.87 & 52.17 \\
Chameleon-7B & 42.19 & 56.25 & 45.31 & 51.56 & 43.75 & 67.19 \\
Qwen2-VL-7B & 47.89 & 47.89 & 56.34 & 49.30 & 47.89 & 46.48 \\
Qwen2.5-VL-7B &67.61	&67.61	&73.24	&39.44	&54.93	&69.01 \\
Molmo-7B & 54.93 & 49.30 & 64.79 & 46.48 & 33.80 & 50.70 \\
MiniCPM-V2-5 & 52.11 & 54.93 & 59.15 & 49.30 & 64.79 & 53.52 \\
MiniCPM-V2-6 & 56.60 & 60.38 & 64.15 & 58.49 & 50.94 & 60.38 \\
InternVL2-8B & 53.52 & 59.15 & 50.70 & 57.75 & 57.75 & 56.34 \\
InternVL2\_5-8B &59.15	&53.52	&63.38	&50.70	&53.52	&57.75 \\
InternVL3-8B &61.97	&60.56	&67.61	&50.70	&47.89	&54.93 \\
GLM-4v & 54.93 & 57.75 & 69.01 & 43.66 & 43.66 & 52.11 \\
Pixtral-12B & 47.89 & 59.15 & 59.15 & 61.97 & 60.56 & 54.93 \\
LLaVA-1.5-13B & 43.28 & 62.69 & 58.21 & 55.22 & 59.70 & 56.72 \\
InternVL-Chat-V1-5 & 52.11 & 54.93 & 57.75 & 57.75 & 49.30 & 47.89 \\
InternVL2-40B & 50.70 & 64.79 & 67.61 & 66.20 & 43.66 & 52.11 \\
Qwen2-VL-72B & 56.34 & 66.20 & 73.24 & 73.24 & 69.01 & 60.56 \\ \hline
\rowcolor{orange! 10}\multicolumn{7}{c}{Novel Baselines} \\ 
Qwen2-VL-SFT & 53.52 & 54.93 & 54.93 & 47.89 & 54.93 & 54.93 \\
Qwen2-VL-DPO & 56.34 & 56.34 & 60.56 & 52.11 & 57.75 & 57.75 \\ \hline
\rowcolor{green! 10}\multicolumn{7}{c}{Medical-Specific Models} \\ 
Med-Flamingo & 62.16 & 45.95 & 43.24 & 43.24 & 45.95 & 48.65 \\
LLava-Med & 47.89 & 50.70 & 53.52 & 56.34 & 47.89 & 47.89 \\
STLLava-Med & 48.65 & 45.95 & 43.24 & 43.24 & 45.95 & 62.16 \\
MedDr & 49.30 & 57.75 & 52.11 & 61.97 & 49.30 & 50.70 \\
HuatuoGPT & 55.71 & 45.71 & 54.29 & 51.43 & 60.00 & 52.86 \\ \hline
\rowcolor{pink! 20}\multicolumn{7}{c}{Proprietary MLLMs} \\
GPT-4o & 53.52 & 56.34 & 61.97 & 66.20 & 54.93 & 61.97 \\
Claude-3.5 & 52.11 & 57.75 & 69.01 & 57.75 & 61.97 & 59.15 \\
Gemini-1.5 &54.42	&59.63	&70.88	&59.01	&63.05	&60.4  \\
O1 &66.27	&67.83	&69.92	&68.10 	&69.14	&71.5 \\
\hline
\end{tabular}}
\caption{Oral\_cavity}
\label{tab:Oral}
\end{table}

\begin{table}[htbp]
\scalebox{0.6}{
\begin{tabular}{l|llllll}
\hline
Model & ACC & REL & COM & CRE & RES & OVE \\ \hline
\rowcolor{gray! 10}\multicolumn{7}{c}{Open-Source MLLMs} \\
VILA1.5-3b &59.26 	&53.70 	&51.85 	&50.00 	&57.41 	&59.26 \\ 
xGen-MM-instruct  &49.02 	&56.86 	&43.14 	&52.94 	&62.75 	&52.94 \\
Deepseek-vl2 & 64.81 & 62.96 & 57.41 & 51.85 & 64.81 & 66.67 \\
Phi-3-5-Vision & 59.26 & 55.56 & 64.81 & 35.19 & 57.41 & 64.81 \\
LLaVA-1.5-7B & 77.36 & 45.28 & 58.49 & 45.28 & 62.26 & 50.94 \\
LLaVA-1.6-7B & 64.81 & 53.70 & 51.85 & 38.89 & 57.41 & 61.11 \\
LLaVA-OneVision-7B & 51.85 & 48.15 & 57.41 & 44.44 & 44.44 & 53.70 \\
Chameleon-7B & 50.00 & 52.08 & 52.08 & 45.83 & 45.83 & 58.33 \\
Qwen2-VL-7B & 59.26 & 53.70 & 59.26 & 50.00 & 61.11 & 57.41 \\
Qwen2.5-VL-7B &66.67	&62.96	&74.07	&53.70	&51.85	&68.52 \\
Molmo-7B & 62.26 & 58.49 & 60.38 & 64.15 & 52.83 & 56.60 \\
MiniCPM-V2-5 & 60.38 & 56.60 & 62.26 & 54.72 & 67.92 & 64.15 \\
MiniCPM-V2-6 & 59.09 & 54.55 & 61.36 & 52.27 & 45.45 & 50.00 \\
InternVL2-8B & 59.26 & 48.15 & 62.96 & 55.56 & 48.15 & 57.41 \\
InternVL2\_5-8B &57.41	&53.70	&50.00	&53.70	&46.30	&55.56 \\
InternVL3-8B &64.81	&68.52	&72.22	&46.30	&57.41	&61.11 \\
GLM-4v & 64.81 & 57.41 & 72.22 & 31.48 & 48.15 & 62.96 \\
Pixtral-12B & 68.52 & 57.41 & 66.67 & 66.67 & 57.41 & 66.67 \\
LLaVA-1.5-13B & 53.85 & 59.62 & 59.62 & 48.08 & 44.23 & 61.54 \\
InternVL-Chat-V1-5 & 68.52 & 42.59 & 53.70 & 64.81 & 61.11 & 64.81 \\
InternVL2-40B & 55.56 & 64.81 & 72.22 & 50.00 & 50.00 & 61.11 \\
Qwen2-VL-72B & 72.22 & 74.07 & 74.07 & 70.37 & 74.07 & 72.22 \\ \hline
\rowcolor{orange! 10}\multicolumn{7}{c}{Novel Baselines} \\ 
Qwen2-VL-SFT & 59.26 & 57.41 & 62.96 & 64.81 & 57.41 & 59.26 \\
Qwen2-VL-DPO & 53.70 & 59.26 & 59.26 & 53.70 & 51.85 & 50.00 \\ \hline
\rowcolor{green! 10}\multicolumn{7}{c}{Medical-Specific Models} \\ 
Med-Flamingo & 50.94 & 62.26 & 58.49 & 45.28 & 45.28 & 47.36 \\
LLava-Med & 77.36 & 45.28 & 58.49 & 45.28 & 62.26 & 50.94 \\
STLLava-Med & 77.36 & 45.28 & 58.49 & 45.28 & 62.26 & 50.94 \\
MedDr & 59.26 & 57.41 & 57.41 & 74.07 & 59.26 & 62.96 \\
HuatuoGPT & 60.38 & 56.60 & 60.38 & 37.74 & 56.60 & 49.06 \\ \hline
\rowcolor{pink! 20}\multicolumn{7}{c}{Proprietary MLLMs} \\
GPT-4o & 59.26 & 68.52 & 55.56 & 64.81 & 57.41 & 59.26 \\
Claude-3.5 & 61.11 & 59.26 & 62.96 & 51.85 & 50.00 & 64.81 \\
Gemini-1.5 &62.79	&60.45	&64.35	&53.22	&51.63	&66.38  \\
O1 &68.04	&66.49	&67.77	&65.35	&65.10 	&69.86 \\
\hline
\end{tabular}}
\caption{Pelvic\_cavity }
\label{tab:Pelvic}
\end{table}

\begin{table}[htbp]
\scalebox{0.6}{
\begin{tabular}{l|llllll}
\hline
Model & ACC & REL & COM & CRE & RES & OVE \\ \hline
\rowcolor{gray! 10}\multicolumn{7}{c}{Open-Source MLLMs} \\
VILA1.5-3b &52.94 	&43.66 	&46.05 	&54.00 	&58.82 	&52.11 \\ 
xGen-MM-instruct  &58.82 	&46.05 	&43.66 	&52.11 	&54.00 	&52.94 \\
Deepseek-vl2 & 42.00 & 56.00 & 54.00 & 52.00 & 46.00 & 42.00 \\
Phi-3-5-Vision & 64.00 & 52.00 & 66.00 & 44.00 & 64.00 & 66.00 \\
LLaVA-1.5-7B & 57.14 & 53.06 & 46.94 & 51.02 & 48.98 & 55.10 \\
LLaVA-1.6-7B & 58.00 & 52.00 & 66.00 & 34.00 & 62.00 & 56.00 \\
LLaVA-OneVision-7B & 59.18 & 59.18 & 57.14 & 44.90 & 55.10 & 55.10 \\
Chameleon-7B & 44.44 & 62.22 & 51.11 & 51.11 & 51.11 & 55.56 \\
Qwen2-VL-7B & 40.00 & 40.00 & 42.00 & 52.00 & 34.00 & 40.00 \\
Qwen2.5-VL-7B &54.00	&48.00	&70.00	&38.00	&58.00	&56.0 \\
Molmo-7B & 54.00 & 52.00 & 60.00 & 40.00 & 60.00 & 50.00 \\
MiniCPM-V2-5 & 67.35 & 40.82 & 67.35 & 44.90 & 61.22 & 61.22 \\
MiniCPM-V2-6 & 56.82 & 59.09 & 65.91 & 54.55 & 47.73 & 59.09 \\
InternVL2-8B & 56.00 & 48.00 & 60.00 & 54.00 & 42.00 & 58.00 \\
InternVL2\_5-8B &50.00	&68.00	&66.00	&56.00	&50.00	&52.00 \\
InternVL3-8B &62.00	&54.00	&72.00	&44.00	&60.00	&68.00 \\
GLM-4v & 56.00 & 56.00 & 68.00 & 34.00 & 60.00 & 58.00 \\
Pixtral-12B & 56.00 & 58.00 & 58.00 & 48.00 & 52.00 & 56.00 \\
LLaVA-1.5-13B & 42.55 & 53.19 & 48.94 & 48.94 & 48.94 & 51.06 \\
InternVL-Chat-V1-5 & 64.00 & 50.00 & 58.00 & 58.00 & 42.00 & 68.00 \\
InternVL2-40B & 60.00 & 46.00 & 52.00 & 52.00 & 54.00 & 60.00 \\
Qwen2-VL-72B & 56.00 & 50.00 & 74.00 & 76.00 & 54.00 & 60.00 \\ \hline
\rowcolor{orange! 10}\multicolumn{7}{c}{Novel Baselines} \\ 
Qwen2-VL-SFT & 50.00 & 60.00 & 60.00 & 54.00 & 60.00 & 54.00 \\
Qwen2-VL-DPO & 58.00 & 48.00 & 60.00 & 52.00 & 52.00 & 56.00 \\ \hline
\rowcolor{green! 10}\multicolumn{7}{c}{Medical-Specific Models} \\ 
Med-Flamingo & 55.10 & 48.98 & 46.94 & 51.02 & 53.06 & 47.14 \\
LLava-Med & 57.14 & 53.06 & 46.94 & 51.02 & 48.98 & 55.10 \\
STLLava-Med & 57.14 & 53.06 & 46.94 & 51.02 & 48.98 & 55.10 \\
MedDr & 60.00 & 56.00 & 58.00 & 68.00 & 56.00 & 60.00 \\
HuatuoGPT & 56.00 & 60.00 & 70.00 & 58.00 & 54.00 & 56.00 \\ \hline
\rowcolor{pink! 20}\multicolumn{7}{c}{Proprietary MLLMs} \\
GPT-4o & 64.00 & 52.00 & 66.00 & 54.00 & 64.00 & 56.00 \\
Claude-3.5 & 60.00 & 46.00 & 52.00 & 52.00 & 54.00 & 60.00 \\
Gemini-1.5 &61.00 	&48.00 	&54.00 	&53.00 	&56.00 	&61.00 \\
O1 &68.00 	&65.00 	&66.00 	&66.00 	&66.00 	&68.00 \\ 
\hline
\end{tabular}}
\caption{Upper\_limb}
\label{tab:Upper}
\end{table}

\begin{table*}[htbp]
\centering
\footnotesize
\renewcommand\arraystretch{1.2}
\tabcolsep=0.1cm
\scalebox{0.85}{
\begin{tabular}{l|llllllll}
\hline
Model &OPH &RAD & ENT & GS  &GI &PULM &CARD &NEURO  \\\hline
\rowcolor{gray! 10}\multicolumn{9}{c}{Open-Source MLLMs} \\
VILA1.5-3b &45.21 &58.57 &56.25 &52.17 &58.23 &59.95 & 45.95 &58.00  \\ 
xGen-MM-instruct &45.21 &59.14 &59.35 &53.26 &56.96 &61.92 &45.95 &60.00  \\
Deepseek-vl2 &57.53 &52.97 &53.66 &54.13
&59.49 &53.69 &56.76 &63.75
 \\
Phi-3-5-Vision &49.32 &61.75 &56.10 &59.43 &60.27
&68.32 &48.65 &70.00 \\
LLaVA-1.5-7B &54.17 &53.79 &63.16 &46.79 &49.37
&53.57 &51.35 &70.00 \\
LLaVA-1.6-7B &46.58 &59.16 &48.78 &55.05 &59.49
&68.97 &45.95 &67.50 \\
LLaVA-OneVision-7B &47.14 &57.00 &46.34 &58.82
&65.38 &60.10 &43.24 &62.03 \\
Chameleon-7B &53.85 &57.76 &56.41 &57.00 &52.11
&60.64 &50.00 &60.87 \\
Qwen2-VL-7B& 54.70 &51.22 &44.04 &65.82 &57.64
&59.46 &51.25 &52.50
\\
Qwen2.5-VL-7B &60.27 &60.64 &58.54 &57.80 &60.76 &59.61 &48.65 &61.25 \\
Molmo-7B &47.89 &52.27 &42.50 &50.46 &53.16 &61.88
&48.65 &52.50\\
MiniCPM-V2-5 &44.44 &52.53 &42.11 &56.88 &54.43 &55.61 &59.46 &50.00
 \\
MiniCPM-V2-6 &53.73 &46.24 &52.63 &57.73 &47.22
&36.51 &52.78 &54.79
 \\
InternVL2-8B &60.27 &56.93 &53.66 &55.96 &55.70
&52.22 &43.24 &57.50 \\
InternVL2\_5-8B &53.42 &55.69 &43.90 &52.29 &63.29 &68.47 &54.05 &68.75 \\
InternVL3-8B &56.16 &61.63 &46.34 &56.88 &48.10 &59.61 &59.46 &57.50 \\
GLM-4v &46.58 &55.97 &43.90 &44.44 &68.42
&55.17 &51.35 &52.50
 \\
LLama-3.2-Vision &55.38 &49.43 &41.03
&48.00 &59.15 &53.72 &35.29 &53.62
 \\
Pixtral-12B &57.53 &61.14 &41.46 &55.96 &65.82
&58.62 &51.35 &67.50 \\
LLaVA-1.5-13B &58.21 &53.49 &47.37 &53.61 &58.33
&54.50 &55.56 &57.53 \\
InternVL-Chat-V1-5 &53.42 &51.74 &58.54 &54.63
&42.11 &54.68 &43.24 &52.50 \\
InternVL2-40B & 53.42 &60.89 &43.90 &55.96 &62.03
&62.07 &45.95 &58.75 \\
Qwen2-VL-72B &64.38 &61.88 &53.66 &57.80 &64.56
&56.16 &54.05 &60.00 \\\hline
\rowcolor{green! 10}\multicolumn{9}{c}{Medical-Specific Models} \\
Med-Flamingo &49.30&56.31&32.50&50.46&53.16
&59.90&62.16&48.75
 \\
LLava-Med &53.52 &53.54 &50.00 &62.39
&53.16 &62.87 &45.95 &56.25
 \\
STLLava-Med &53.42 &57.43 &56.10 &51.38
&67.09 &50.74 &48.65 &62.50
 \\
MedDr &56.16 &59.25 &63.41 &54.72
&71.23 &55.94 &54.05 &65.00\\
HuatuoGPT-Vision &52.05 &58.42 &46.34 &58.72
&58.23 &68.97 &51.35 &65.00
\\ \hline
\rowcolor{pink! 20}\multicolumn{9}{c}{Proprietary MLLMs} \\
GPT-4o &57.53 &55.71 &59.38 &57.61 &48.10 & 55.67 &45.95 &61.25 \\
Claude-1.5 &60.27 &56.86 &50.00 &59.78 &53.16 &58.62 &51.35 &52.50 \\
Gemini-1.5 &60.27 &61.39 &68.29 &57.80 &74.68 &57.14
&62.16 &73.75 \\
O1 &68.49 &65.59 &58.54 &62.39 &75.95
&59.61 &70.27 &70.00
\\\hline
\end{tabular}}
\caption{The overall performance of different MLLMs in judging, compared with human annotations in different departments. OPH, RAD, ENT, GS, GI, PULM, CARD, and NEURO denote Ophthalmology, Radiology, Otolaryngology, General Surgery, Gastroenterology, Pulmonology, Cardiology,and Neurology, respectively.}
\label{tab:department1}
\end{table*}



\end{document}